\documentclass[11pt]{article}

\usepackage[preprint]{acl}

\makeatletter
\ifdefined\@LN@makecol
  \IfFormatAtLeastTF{2025-06-01}{%
    \AtBeginDocument{%
      \let\@LN@orig@makecol\@makecol
      \def\@LN@makecol{%
        \@LN@orig@makecol
        \setbox\@outputbox\vbox{%
          \boxmaxdepth\@maxdepth
          \protected@write\@auxout{}{%
            \string\def\string\@LN@column{\if@firstcolumn1\else2\fi}%
          }%
          \box\@outputbox
        }%
      }%
      \RemoveFromHook{build/column/before}[lineno]%
    }%
  }{}%
\fi
\makeatother

\usepackage{times}
\usepackage{latexsym}
\usepackage[T1]{fontenc}
\usepackage[utf8]{inputenc}
\usepackage{microtype}
\usepackage{inconsolata}
\usepackage{graphicx}

\usepackage{amsmath,amsfonts,bm}

\def\eqref#1{equation~\ref{#1}}

\def\1{\bm{1}}

\DeclareMathAlphabet{\mathsfit}{\encodingdefault}{\sfdefault}{m}{sl}
\SetMathAlphabet{\mathsfit}{bold}{\encodingdefault}{\sfdefault}{bx}{n}

\DeclareMathOperator{\sign}{sign}

\usepackage{amsmath}
\usepackage{amsthm}
\usepackage{booktabs}
\usepackage{multirow}
\usepackage{xcolor}
\usepackage{colortbl}
\definecolor{bsdrow}{gray}{0.90}
\definecolor{bsdblue}{HTML}{0072B2}
\definecolor{wrongred}{HTML}{C0392B}
\usepackage{pifont}

\newtheorem{proposition}{Proposition}
\DeclareMathOperator{\sg}{sg}
\DeclareMathOperator{\clip}{clip}
\newcommand{\clipe}[1]{\clip_{1\pm\epsilon}\!\left(#1\right)}

\title{Bridged Self-Distillation for On-Policy Learning}

\newcommand{\affil}[1]{\textsuperscript{\normalfont #1}}
\author{
  Runsheng Chen\affil{1,*},
  Yihong Liu\affil{2,3,*,\dag},
  Raoyuan Zhao\affil{2,3}, \\
  \bfseries Michael A. Hedderich\affil{2,3},
  Hinrich Sch{\"u}tze\affil{2,3},
  and Matthias Grabmair\affil{1} \\[4pt]
  \textsuperscript{1}Technical University of Munich \quad
  \textsuperscript{2}LMU Munich \\
  \textsuperscript{3}Munich Center for Machine Learning (MCML) \\
  \texttt{\{runsheng.chen, matthias.grabmair\}@tum.de} \\
  \texttt{\{yihong, rzhao, hedderich\}@cis.lmu.de}
}

\begin{document}
\maketitle
{\renewcommand{\thefootnote}{*}\footnotetext{Equal contribution.}%
 \renewcommand{\thefootnote}{\dag}\footnotetext{Corresponding author.}}

\begin{abstract}
On-policy distillation can transfer the reasoning capabilities of a strong teacher to a smaller student.
However, distillation from a stronger teacher model faces two mismatches: its solutions may be difficult for a smaller student to imitate, and its token-level probabilities may be incompatible with, or even unavailable to, the student. 
Existing methods handle these settings through different forms of supervision.
To this end, we propose \emph{\textbf{B}ridged \textbf{S}elf-\textbf{D}istillation} (\textbf{BSD}), a unified framework for distilling from \emph{any} teacher that provides reasoning traces.
The model to be distilled from, termed the \emph{external teacher}, first provides a reasoning trace as privileged context, which can be generated in an offline fashion.
The student generates rollouts without this context, while a copy of the student conditioned on it, serving as the \emph{self-teacher}, scores the same rollouts in the student's own token space.
BSD uses the resulting token-level guidance to adjust updates on unsuccessful rollouts, with a verifier providing the outcome signal. 
As the external teacher is queried only to produce text, the same training procedure applies to all teacher scenarios: shared-tokenizer, cross-tokenizer, and closed API teachers.
Across these scenarios, BSD improves over the base model and outperforms baselines tailored to each scenario, offering a novel and unified pathway for distillation. Code is available: \url{https://github.com/TUMLegalTech/Bridged-Self-Distillation}
\end{abstract}

\begin{figure}[t]
  \centering
    \setlength{\belowcaptionskip}{-0.3cm}
  \includegraphics[width=0.42\textwidth]{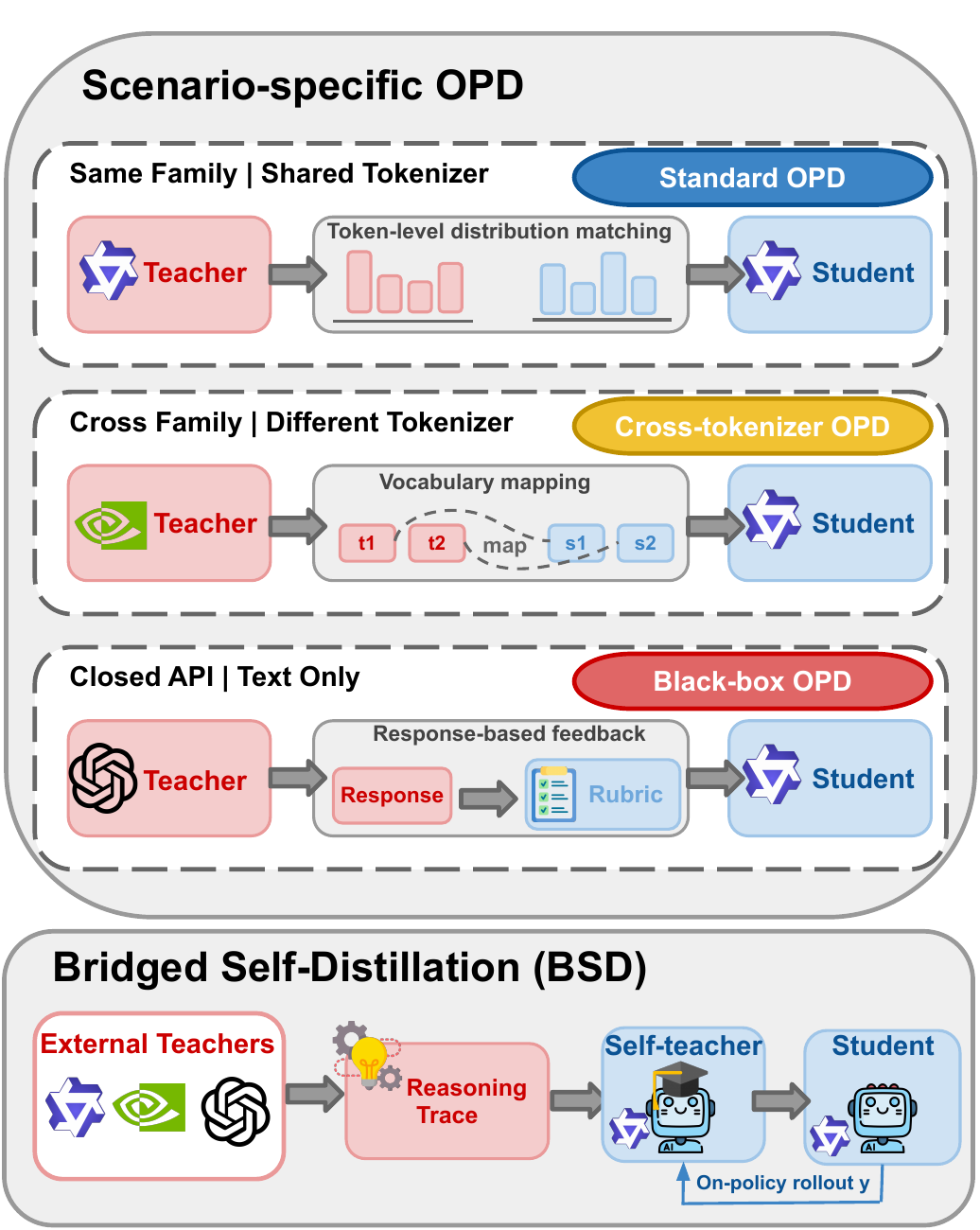}
  \caption{Comparison between scenario-specific OPD and BSD.
  \textbf{Top}: existing methods derive a different training signal for each kind of teacher access, namely token-level distribution matching with a shared tokenizer, vocabulary mapping across tokenizers, and response-based feedback for a closed API.
  \textbf{Bottom}: BSD uses a unified framework for all three scenarios: the external teacher
  provides a reasoning trace $z$, and a self-teacher conditioned on $z$ scores the student's own on-policy rollouts.}
  \label{fig:teaser}
\end{figure}

\section{Introduction}
\label{sec:intro}

Large reasoning models have become increasingly capable on tasks that require multi-step reasoning \citep{wei2022cot,ahn-etal-2024-large,muennighoff-etal-2025-s1,deepseekr1}. 
Much of this progress comes from post-training, typically including supervised fine-tuning (SFT) on reasoning traces and reinforcement learning with verifiable rewards (RLVR) \citep{ouyang2022rlhf,Chung2024sft,shao2024deepseekmath,ye-etal-2025-analyzing,ye2025limoreasoning,yu2025dapo}.
Yet the cost of running a large reasoning model makes it desirable to transfer its capabilities to smaller student models that can respond more quickly and be deployed more widely \citep{hinton2015distilling,kim2016sequence}. 
On-policy distillation (OPD) has emerged as a promising post-training approach for this purpose, allowing a smaller model to learn from the guidance of a stronger teacher \citep{gu2024minillm,agarwal2024gkd,ko2024distillm,xu2025skd,lu2025onpolicyblog,li2026rethinkingopd,song2026opdsurvey}.

In standard OPD, the student generates its own rollouts, and the teacher scores the student's tokens to guide training \citep{ross2011dagger,agarwal2024gkd,jin2026entropyopd,wu2026lightningopd}.
This requires access to the teacher's next-token distributions and a way to align them with the student's tokenization.
However, when teacher and student use different tokenizers, these distributions cannot be compared directly; when the teacher is available only through a closed API, they may be inaccessible.
\emph{Cross-tokenizer} OPD methods address the former by constructing mappings between the token spaces of the student and teacher \citep{zhang-etal-2024-dual,minixhofer2025universal,boizard2025crosstokenizerdistillationuniversallogit,niu2026breaking,sun2026simctrecoveringlostsupervision}, while \emph{black-box} methods address the latter using feedback such as rubrics or sampled continuations \citep{jiang-etal-2023-lion,ye2026blackboxonpolicydistillationlarge,fang2026ropd,xiong2026ovdonpolicyverbaldistillation,zhou2026omniopd}.
These methods expand the range of usable teachers, but require different training signals for different teacher settings (Figure~\ref{fig:teaser}, top).
Alternatively, SFT can train on teacher-generated text without accessing probabilities, although directly imitating a much stronger model's reasoning can be difficult for a small student \citep{zhang-etal-2023-lifting,zhang-etal-2025-towards-law,huang2026finetunereasoningmodelteacherstudent,lyu2026studentcentereddistillationnarrowsagentic}. 
This motivates \emph{a unified on-policy learning method that can distill from any teacher capable of providing reasoning traces}.

We propose \emph{\textbf{B}ridged \textbf{S}elf-\textbf{D}istillation} (\textbf{BSD}) to turn the reasoning of an \emph{external teacher} into guidance on the student's own rollouts (Figure~\ref{fig:teaser}, bottom).
The external teacher is the model we aim to distill from, and BSD requires only its reasoning trace.
Instead of training the student to imitate the teacher's solution, as in SFT, we provide that solution as privileged context to a copy of the student.
Inspired by on-policy self-distillation \citep{zhao2026opsd,zhang2026opsdlonpolicyselfdistillationlongcontext,yang2026selfdistilled}, this conditioned copy serves as a \emph{self-teacher}, scoring rollouts generated by the student 
without the privileged context.
Both versions score the same tokens in the student's own token space, and the differences show how access to the teacher's reasoning changes the student's predictions along its rollouts.
BSD uses these differences to adjust token-level penalties on unsuccessful rollouts, while a verifier determines the direction of the updates.
Because the external teacher provides only text, the same training procedure applies to shared-tokenizer, cross-tokenizer, and closed API teachers.

We evaluate BSD under three forms of teacher access: (1) shared-tokenizer, (2) cross-tokenizer, and (3) closed API teachers.
Across twelve teacher--student pairs and three math reasoning benchmarks, BSD remains effective in all three settings while avoiding the severe degradation that SFT exhibits in several pairs.
Our contributions are as follows:

\begin{itemize}
    \item We introduce BSD, a unified on-policy distillation framework that turns reasoning traces from external teachers into token-level guidance through a privileged self-teacher.
    \item We evaluate BSD across twelve teacher--student pairs spanning shared-tokenizer, cross-tokenizer, and closed API teachers, demonstrating that BSD generally outperforms the baselines across all three settings.
    \item Our analyses show that BSD maintains stable training, benefits from more complete reasoning traces, and concentrates its guidance on uncertain tokens, with stronger guidance associated with larger performance gains. 
\end{itemize}

\section{Preliminaries}
\label{sec:setup}

\textbf{Group Relative Policy Optimization (GRPO).}

Let $\pi_\theta$ denote the policy model being optimized.
For a question $x$, the policy samples a group of rollouts $\{y^{(i)}\}_{i=1}^{G}$, and a verifier assigns each rollout a binary reward $r^{(i)}\in\{0,1\}$.
GRPO~\citep{shao2024deepseekmath} normalizes these rewards within the group:
\setlength{\abovedisplayskip}{3pt}
\setlength{\belowdisplayskip}{3pt}
\begin{equation}
  A^{(i)} \;=\;
  \frac{r^{(i)} - \operatorname{mean}\big(\{r^{(j)}\}_{j=1}^{G}\big)}
       {\operatorname{std}\big(\{r^{(j)}\}_{j=1}^{G}\big)},
  \label{eq:grpo}
\end{equation}
and applies the advantage $A^{(i)}$ uniformly to every token of $y^{(i)}$
in a clipped policy-gradient update.
When a group contains both outcomes, successful rollouts have positive advantage and unsuccessful rollouts have negative advantage.
If all rollouts have the same outcome, their advantages are zero, and the group contributes no gradient.
We omit the rollout index $i$ when considering a single rollout.

\begin{figure*}[t]
  \centering
    \setlength{\belowcaptionskip}{-0.2cm}
  \includegraphics[width=0.95\textwidth]{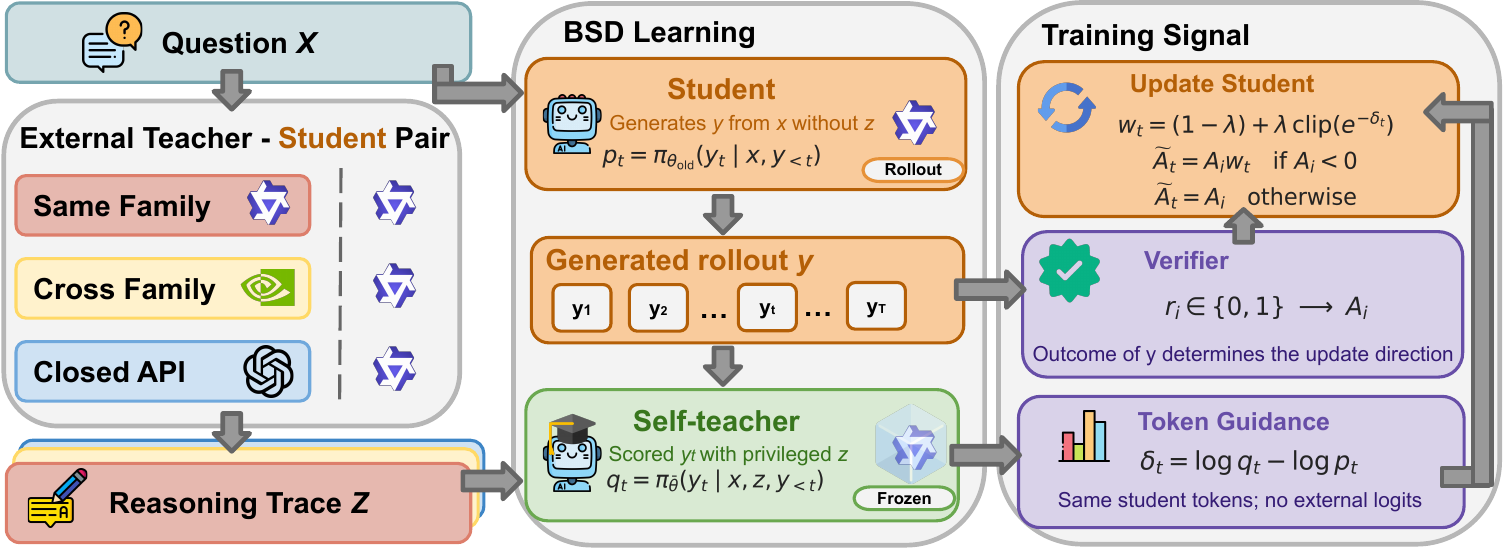}
  \caption{Overview of Bridged Self-Distillation (BSD). An external teacher, whatever its access
  setting, provides a reasoning trace $z$ for each question. 
  The student generates rollouts without $z$, and the self-teacher (a frozen copy of the student conditioned on $z$) scores the same tokens.
  The token-level guidance $\delta_t$ reweights the penalty on unsuccessful rollouts, while the verifier determines the direction of
  every update.}
  \label{fig:pipeline}
\end{figure*}

\textbf{Privileged self-distillation.}
A privileged context $z$ is available during training but absent when the model generates rollouts or performs inference.
In on-policy self-distillation (OPSD)~\citep{zhao2026opsd,huebotter2026sdpo}, the model generates on-policy rollouts without $z$ as the student.
A copy conditioned on $z$ acts as the \emph{teacher} and scores the same tokens.
RLSD~\citep{yang2026selfdistilled} uses the ratio of these two token probabilities to refine the rollout-level advantage at each token:
\begin{equation}
\begin{aligned}
  u_t &=
  \left[
    \frac{\pi_\theta(y_t\mid x,z,y_{<t})}
         {\pi_\theta(y_t\mid x,y_{<t})}
  \right]^{\sign(A)},\\
  \hat A_t^{\mathrm{RLSD}}
  &= A\big[(1-\lambda)+\lambda\,\clipe{u_t}\big].
\end{aligned}
\label{eq:rlsd}
\end{equation}
Here, $\clipe{\cdot}$ clips its argument to $[1-\epsilon,1+\epsilon]$, and $\lambda\in[0,1]$ controls the strength of token-level reweighting.
Because the multiplier is positive, reweighting preserves the sign of $A$.
Intuitively, in RLSD, the verifier determines the direction of the update while the privileged context changes how strongly individual tokens are updated.

\section{Bridged Self-Distillation}
\label{sec:method}

To distill from teacher models with different tokenizers or text-only access, we need guidance that does not depend on their token probabilities.
Directly imitating a stronger teacher's reasoning can also be difficult for a smaller student.
We therefore propose \emph{\textbf{B}ridged \textbf{S}elf-\textbf{D}istillation} (\textbf{BSD}), which only needs a teacher's reasoning trace and turns it into guidance on the student's own rollouts (Figure~\ref{fig:pipeline}).
We describe the key components below.

\subsection{External Teacher}
\label{sec:bridge}

We refer to the model that we want to distill from as the \emph{external teacher}.
Given a question $x$, it provides a \emph{reasoning trace} $z$, the text it returns when solving $x$, which contains its full reasoning when this is exposed and only its final response otherwise.
BSD uses $z$ as the privileged context for the \emph{self-teacher}. 
Because the external teacher only needs to provide text, this setup applies to all scenarios, including shared-tokenizer, cross-tokenizer, and closed API teachers.

\subsection{Self-Teacher}
\label{sec:selfteacher}

To turn $z$ into guidance on the student's own reasoning, the student first generates a rollout $y$ from $x$ without $z$.
For each generated token $y_t$, the student assigns the probability $p_t = \pi_\theta(y_t \mid x, y_{<t})$.

The \emph{self-teacher} is a frozen copy $\pi_{\bar\theta}$ of the student conditioned on both $x$ and $z$.
Rather than generating a separate solution, the self-teacher scores the same rollout $y$ token by token, assigning $q_t = \pi_{\bar\theta}(y_t \mid x, z, y_{<t})$.
As the two scores come from the same model and share the same token space, neither
logits from the external teacher nor vocabulary alignment is needed.

We define the token-level guidance as their per-token log-probability difference:
\begin{equation}
  \delta_t \;=\; \sg\big[\log q_t - \log p_t\big],
  \label{eq:delta}
\end{equation}
where $\sg[\cdot]$ denotes the stop-gradient operator.
A positive $\delta_t$ means the self-teacher assigns more probability to the student's chosen token after reading $z$; a negative value means it assigns less.

\subsection{Learning}
\label{sec:bsd}

The self-teacher provides token-level guidance, while a verifier checks whether each rollout reaches the correct answer.
GRPO converts these outcomes into the group-relative advantage $A$ (Eq.~\ref{eq:grpo}).
For an unsuccessful rollout with $A<0$, BSD instead uses the guidance $\delta_t$ to set a token-specific weight:
\begin{equation}
  w_t \;=\; (1-\lambda) + \lambda\,\clipe{e^{-\delta_t}}.
  \label{eq:weight}
\end{equation}
When $\delta_t<0$, the self-teacher assigns less probability to the chosen token, so $w_t>1$ and that token receives a stronger penalty.
When $\delta_t>0$, the penalty is reduced.
The coefficient $\lambda$ controls the strength of this reweighting, which decays
over training, while clipping bounds its effect.

The weight $w_t$ is applied only to unsuccessful rollouts, while the successful ones are unchanged:
\begin{equation}
  \tilde A_t \;=\;
  \begin{cases}
    A\,w_t & \text{if } A < 0, \\
    A & \text{otherwise.}
  \end{cases}
  \label{eq:routing}
\end{equation}
This means that successful rollouts retain their advantage, as in GRPO, allowing the student to reinforce correct solutions it found without the external teacher's guidance.
Because $w_t$ is positive, the token-level guidance changes the magnitude of a penalty without changing the sign of $A$.

The student model is then optimized by maximizing the following policy-gradient objective $\mathcal{J}(\theta) = \mathbb{E}\big[\sum_t \tilde A_t \log \pi_\theta(y_t \mid x, y_{<t})\big]$.\footnote{BSD coincides with GRPO on successful rollouts and, to first order, adds an OPD-style correction on unsuccessful rollouts; see Proposition~\ref{prop:smallgap} in Appendix~\ref{app:proofs}.}

\section{Experimental Setup}
\label{sec:exp_setup}

\subsection{Models}

\textbf{Backbone models.}
We use three Qwen3 models as the student models~\citep{qwen3report} (in thinking mode): \texttt{Qwen3-0.6B}, \texttt{Qwen3-1.7B}, and \texttt{Qwen3-4B}.
We also use \texttt{Qwen3-4B-Instruct-2507} as an additional backbone, which we first fine-tune on 1{,}000 reasoning traces 
to enable it to produce reasoning blocks.
We refer to the fine-tuned model as \texttt{Qwen3-4B-Instruct} and use it as another student.

\textbf{Teacher models.}
We consider three models as external teachers with different access settings: \texttt{Qwen3-32B}~\citep{qwen3report} as a shared-tokenizer teacher, \texttt{Nemotron-3-30B-A3B}~\citep{nemotron3} as a cross-tokenizer teacher, and \texttt{GPT-5.6 Luna}~\citep{gpt56luna}
as a closed API teacher.

\subsection{Training}

\textbf{Data.}
We use the mathematical reasoning subset of OpenThoughts~\citep{guha2025openthoughts}. 
For each teacher--student pair, we sample 2{,}400 training problems, and the external teacher provides one reasoning trace $z$ per problem.
The traces of \texttt{Qwen3-32B} and \texttt{Nemotron-3-30B-A3B} average about 4{,}000 tokens, while \texttt{GPT-5.6 Luna} returns only its final response, which averages 247 tokens.

\textbf{Implementation details.}
We implement BSD using verl~\citep{sheng2025hybridflow} and vLLM~\citep{kwon2023vllm}.
Each student model is trained for 150 updates
on NVIDIA H100/H200 GPUs.
Further details are provided in Appendix~\ref{app:exp_details}.

\subsection{Evaluation}

\textbf{Benchmarks.}
We evaluate on three competition-level mathematics benchmarks: AIME 2024~\citep{aime24}, AIME 2025~\citep{aime25}, and HMMT 2025~\citep{dekoninck2026matharena}.
We select the best checkpoints using a separate set of 128 development problems from OpenThoughts.

\textbf{Metrics.}
For each benchmark problem, we generate four samples.
We report Avg@4, the average correctness of the four samples, in Table~\ref{tab:main}, and Pass@4, the proportion of problems solved by at least one sample, in Table~\ref{tab:pass4}.
The evaluation settings, the teachers' own scores, and the selected checkpoints are in Appendices~\ref{app:exp_details} and~\ref{app:steps}.

\begin{table*}[t]
\centering
\small
\setlength{\belowcaptionskip}{-0.3cm}
\setlength{\tabcolsep}{4pt}
\renewcommand{\arraystretch}{0.80}
\resizebox{\textwidth}{!}{%
\begin{tabular}{l cccc cccc cccc}
\toprule
 & \multicolumn{4}{c}{Shared tokenizer (\texttt{Qwen3-32B})} & \multicolumn{4}{c}{Cross-tokenizer (\texttt{Nemotron-3-30B-A3B})} & \multicolumn{4}{c}{Closed API (\texttt{GPT-5.6 Luna})} \\
\cmidrule(lr){2-5} \cmidrule(lr){6-9} \cmidrule(lr){10-13}
Method & AIME24 & AIME25 & HMMT25 & Avg & AIME24 & AIME25 & HMMT25 & Avg & AIME24 & AIME25 & HMMT25 & Avg \\
\midrule
\multicolumn{13}{l}{\texttt{Qwen3-4B}} \\
Base & 72.5 & 62.5 & 42.5 & 59.2 & 72.5 & 62.5 & 42.5 & 59.2 & 72.5 & 62.5 & 42.5 & 59.2 \\
+ SFT & 72.5 & 65.0 & \textbf{44.2} & 60.6 & \textbf{76.7} & 65.0 & \textbf{49.2} & \textbf{63.6} & 68.3 & 62.5 & 40.8 & 57.2 \\
+ OPD & 73.3 & 67.5 & 40.0 & 60.3 & 70.8 & 62.5 & 45.8 & 59.7 & 70.8 & 64.2 & 44.2 & 59.7 \\
\rowcolor{bsdrow} + BSD & \textbf{79.2} & \textbf{68.3} & \textbf{44.2} & \textbf{63.9} & \textbf{76.7} & \textbf{66.7} & 47.5 & \textbf{63.6} & \textbf{75.8} & \textbf{65.0} & \textbf{45.0} & \textbf{61.9} \\
\midrule
\multicolumn{13}{l}{\texttt{Qwen3-4B-Instruct}} \\
Base & 56.7 & 49.2 & 27.5 & 44.4 & 56.7 & 49.2 & 27.5 & 44.4 & 56.7 & 49.2 & 27.5 & 44.4 \\
+ SFT & 50.0 & 46.7 & 25.8 & 40.8 & 47.5 & 43.3 & 23.3 & 38.1 & 47.5 & 47.5 & 24.2 & 39.7 \\
+ OPD & 54.2 & 45.0 & 27.5 & 42.2 & 54.2 & \textbf{50.8} & 30.0 & 45.0 & 57.5 & \textbf{50.8} & \textbf{29.2} & 45.8 \\
\rowcolor{bsdrow} + BSD & \textbf{57.5} & \textbf{51.7} & \textbf{28.3} & \textbf{45.8} & \textbf{55.8} & 49.2 & \textbf{31.7} & \textbf{45.6} & \textbf{60.0} & \textbf{50.8} & 28.3 & \textbf{46.4} \\
\midrule
\multicolumn{13}{l}{\texttt{Qwen3-1.7B}} \\
Base & 50.8 & 34.2 & 22.5 & 35.8 & 50.8 & 34.2 & 22.5 & 35.8 & 50.8 & 34.2 & 22.5 & 35.8 \\
+ SFT & 38.3 & 31.7 & 20.0 & 30.0 & 53.3 & 34.2 & 25.0 & 37.5 & 55.0 & 37.5 & 18.3 & 36.9 \\
+ OPD & \textbf{56.7} & 34.2 & \textbf{25.0} & \textbf{38.6} & 45.0 & 33.3 & \textbf{26.7} & 35.0 & 49.2 & 35.8 & 24.2 & 36.4 \\
\rowcolor{bsdrow} + BSD & 53.3 & \textbf{37.5} & \textbf{25.0} & \textbf{38.6} & \textbf{55.8} & \textbf{38.3} & 23.3 & \textbf{39.2} & \textbf{56.7} & \textbf{38.3} & \textbf{25.0} & \textbf{40.0} \\
\midrule
\multicolumn{13}{l}{\texttt{Qwen3-0.6B}} \\
Base & 8.3 & 17.5 & 6.7 & 10.8 & 8.3 & 17.5 & 6.7 & 10.8 & 8.3 & 17.5 & 6.7 & 10.8 \\
+ SFT & 7.5 & 13.3 & 6.7 & 9.2 & 5.8 & 10.0 & 6.7 & 7.5 & 0.8 & 5.0 & 2.5 & 2.8 \\
+ OPD & 11.7 & \textbf{17.5} & \textbf{10.0} & 13.1 & \textbf{15.0} & \textbf{20.0} & 7.5 & \textbf{14.2} & \textbf{11.7} & 14.2 & 8.3 & 11.4 \\
\rowcolor{bsdrow} + BSD & \textbf{14.2} & \textbf{17.5} & 9.2 & \textbf{13.6} & 12.5 & 17.5 & \textbf{8.3} & 12.8 & 10.0 & \textbf{18.3} & \textbf{9.2} & \textbf{12.5} \\
\bottomrule
\end{tabular}}
\caption{Avg@4 on three mathematical reasoning benchmarks; Avg is the mean across benchmarks.
OPD denotes vanilla OPD, xOPD, and ROPD for shared-tokenizer, cross-tokenizer, and closed API teachers, respectively.
\textbf{Bold} marks the best score among distillation methods within each teacher--student setting.
BSD consistently improves over the base model in all twelve teacher--student pairs and achieves the best Avg in eleven pairs.
}
\label{tab:main}
\end{table*}

\subsection{Baselines}

We consider the following baselines, which use the same backbones and training problems as BSD.

\textbf{Supervised Fine-Tuning (SFT).}
We train each student to imitate the external teacher's reasoning traces \citep{kim2016sequence}
using the same traces that serve as privileged context for BSD.

\textbf{Vanilla OPD.}
For the shared-tokenizer teacher, we use standard OPD~\citep{lu2025onpolicyblog}, which is optimized by minimizing the log-probability differences between student and teacher.

\textbf{Cross-Tokenizer OPD (xOPD).}
For the cross-tokenizer teacher, we use cross-tokenizer OPD~\citep{niu2026breaking}, denoted xOPD, which aligns the two tokenizations and maps teacher token log-probabilities onto student tokens.

\textbf{Rubric-Based OPD (ROPD).}
For the closed API teacher, we use ROPD~\citep{fang2026ropd}, which asks the teacher to derive a rubric for each question and score the student's rollouts against it.

\section{Main Results}
\label{sec:results}

We examine how BSD performs under different forms of teacher access.
Across twelve teacher--student pairs, we first compare BSD with SFT and the OPD baseline designed for each of the shared-tokenizer, cross-tokenizer, and closed API settings. 
We then track performance and response length throughout training to examine how BSD and the baselines affect the student's behavior.

\subsection{Comparing Teacher Access}
\label{sec:main_results}

\textbf{BSD consistently improves over the base model and outperforms the distillation baselines across teacher settings.}
Table~\ref{tab:main} reports Avg@4 for four students paired with shared-tokenizer, cross-tokenizer, and closed API teachers.
BSD improves over the base model in \textbf{all twelve} teacher--student pairs, with gains of 1.2 to 4.7 points, and achieves the best average score in \textbf{eleven} pairs.

\textbf{Shared-tokenizer teacher.}
\texttt{Qwen3-32B} shares the students' tokenizer, allowing vanilla OPD to score their rollouts directly with teacher token probabilities.
BSD outperforms SFT for all four students by 3.3 to 8.6 points in average score, while SFT falls below the base model for three of them.
Thus, even within the same model family, directly imitating a much stronger teacher's reasoning can be difficult for smaller students~\citep{zhang-etal-2023-lifting,zhang-etal-2025-towards-law}.
BSD also matches or outperforms vanilla OPD for every student: the two tie in average score for \texttt{Qwen3-1.7B}, and BSD scores up to 3.6 points higher for the other three.
This indicates that sharing a tokenizer does not ensure compatible reasoning patterns between distinct models, which prior work identifies as important for successful OPD~\citep{li2026rethinkingopd}.
BSD instead uses a self-teacher, a copy of the student conditioned on the external teacher's reasoning, so that the token-level guidance comes from the student's own model and is therefore more compatible.

\begin{figure*}[t]
  \centering
  \setlength{\belowcaptionskip}{-0.3cm}
  \includegraphics[width=0.95\textwidth]{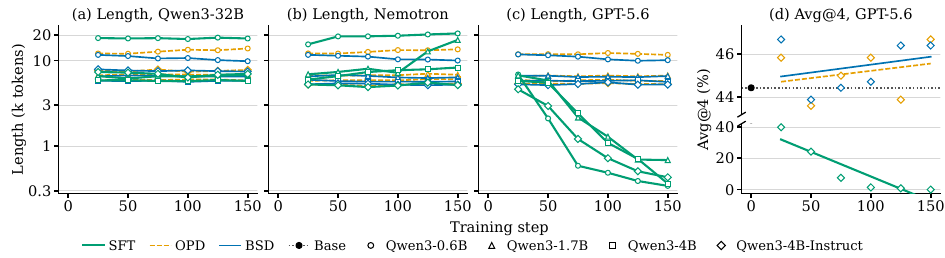}
  \caption{Training progression.
\textbf{(a)--(c)} Mean response length on the development set at each checkpoint for four students under the shared-tokenizer, cross-tokenizer, and closed API teachers, respectively. 
OPD refers to vanilla OPD, xOPD, and ROPD in the three respective settings.
\textbf{(d)} Avg@4 across the three benchmarks at each checkpoint for \texttt{Qwen3-4B-Instruct} distilled from \texttt{GPT-5.6 Luna}.
SFT can substantially shorten its responses and cause performance to collapse with the closed API teacher, whereas BSD and OPD generally maintain stable response lengths and show stable or improving performance throughout training.
}
  \label{fig:progression}
\end{figure*}

\textbf{Cross-tokenizer teacher.}
Because the teacher model \texttt{Nemotron-3-30B-A3B} uses a different tokenizer from the students, xOPD must align the two tokenizations and map the teacher's token probabilities onto the student's tokens.
Nevertheless, xOPD does not yield consistent gains over the base models.
It improves the average score of \texttt{Qwen3-0.6B} by 3.4 points, but brings only marginal gains of 0.5 and 0.6 points for \texttt{Qwen3-4B} and \texttt{Qwen3-4B-Instruct}, respectively, and decreases the score of \texttt{Qwen3-1.7B} by 0.8 points.
This indicates that constructing a mapping makes cross-tokenizer OPD possible, but does not guarantee effective token-level supervision.
In contrast, BSD improves all four base models by 1.2 to 4.4 points and outperforms xOPD for three of the four students,
with \texttt{Qwen3-0.6B} as the only exception.
BSD therefore provides a strong alternative for cross-tokenizer distillation, obtaining guidance directly in the student's token space without mappings between distinct vocabularies.

\textbf{Closed API teacher.}
\texttt{GPT-5.6 Luna} is accessible only through a text API and exposes neither token probabilities nor its tokenizer.
Moreover, it returns short final responses averaging 247 tokens, compared with about 4{,}000 tokens from the other two teachers.
Directly using these responses as SFT targets causes the students to imitate their length (cf.\ Section~\ref{sec:progression}).
Consequently, SFT falls below the base model for three of the four students, with \texttt{Qwen3-0.6B} dropping from 10.8 to 2.8.
ROPD handles the lack of token probabilities by asking the external teacher to derive a rubric and score student rollouts during training.
BSD outperforms SFT for all four students by 3.1 to 9.7 points and ROPD by 0.6 to 3.6 points.
BSD achieves these gains by using each API response as privileged context for the self-teacher while allowing the student to generate its own rollouts, without training the student to reproduce their form.

\subsection{Training Progression}
\label{sec:progression}

We further examine how distillation changes the students throughout training.
Figure~\ref{fig:progression} tracks two aspects of this progression: response length across all twelve teacher--student pairs and benchmark performance in a representative closed API setting.\footnote{More performance curves for other teacher--student pairs are provided in Appendix~\ref{app:dev_curves}, where SFT performance also frequently deteriorates over training.}

\textbf{Response length.}
Figure~\ref{fig:progression}a--c shows that the effect of SFT on response length depends strongly on the external teacher.
With the shared-tokenizer teacher, response lengths remain largely stable for all methods.
With the cross-tokenizer teacher, BSD and xOPD remain stable, whereas SFT usually increases the response lengths, most notably for \texttt{Qwen3-1.7B}.
The most pronounced shift occurs with the closed API teacher, whose responses average only 247 tokens.
SFT trains directly on these short responses, causing all four students to gradually shorten their outputs.
In contrast, BSD and the corresponding OPD baselines maintain relatively stable response lengths across all twelve pairs.
These trends show that using the external teacher's reasoning trace as privileged context allows BSD to preserve the student's output form without forcing it to reproduce the trace's length.

\begin{figure*}[t]
  \centering
  \setlength{\belowcaptionskip}{-0.3cm}
  \begin{minipage}[t]{0.32\textwidth}
    \centering
    (a)
    \includegraphics[width=\linewidth]{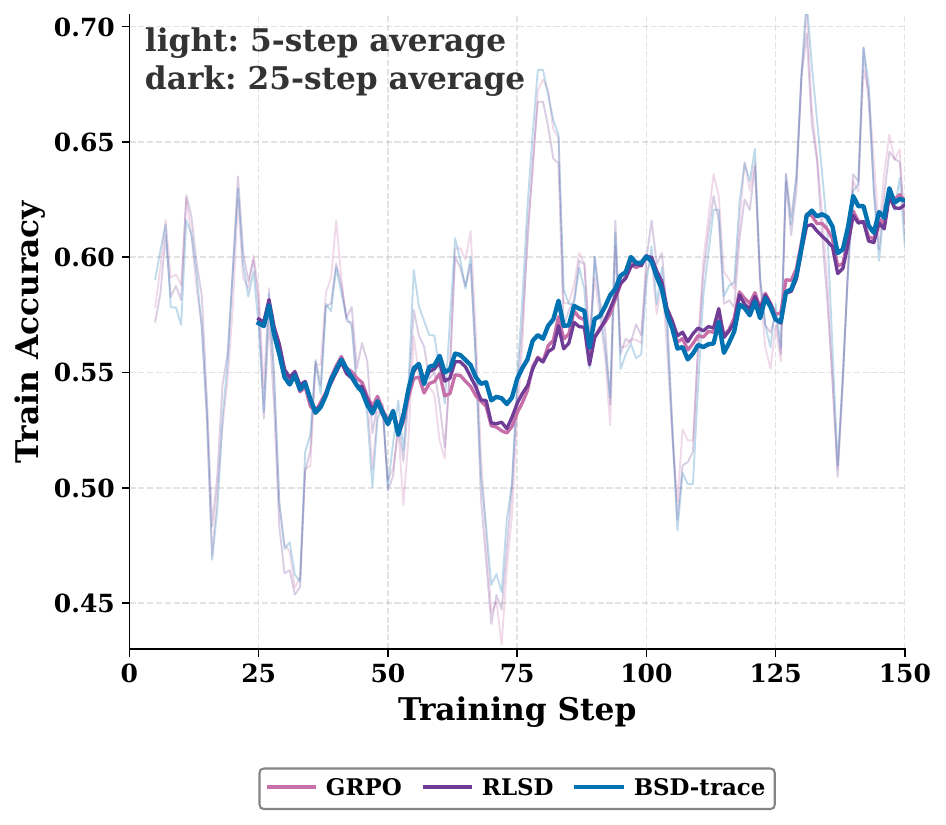}
  \end{minipage}
  \hfill
  \begin{minipage}[t]{0.32\textwidth}
    \centering
    (b)
    \includegraphics[width=\linewidth]{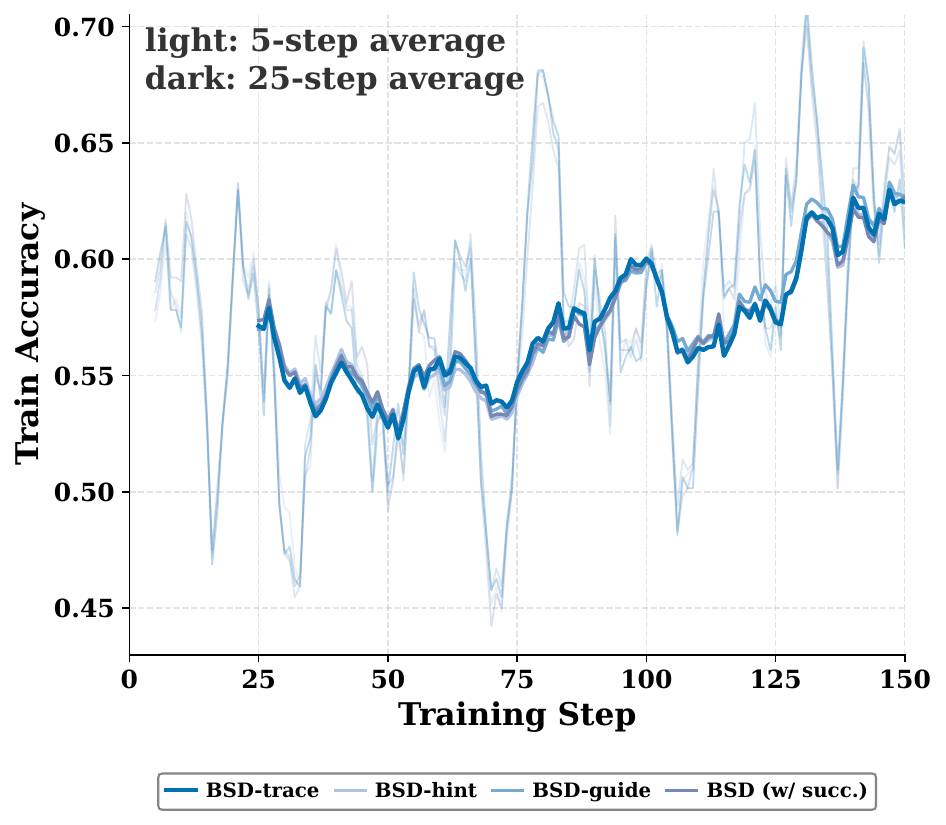}
  \end{minipage}
  \hfill
  \begin{minipage}[t]{0.32\textwidth}
    \centering
    (c)
    \includegraphics[width=\linewidth]{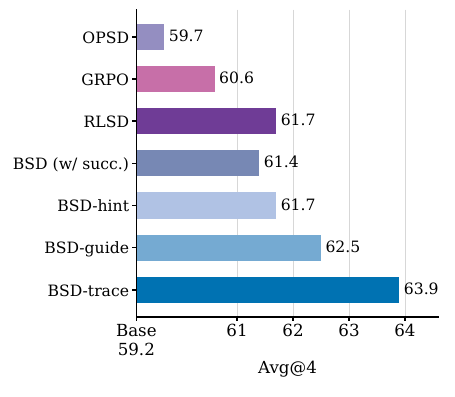}
  \end{minipage}
  \caption{Ablations with the \texttt{Qwen3-4B} student and \texttt{Qwen3-32B} external teacher.
\textbf{(a)} Training accuracy of GRPO, answer-conditioned RLSD, and full-trace BSD.
\textbf{(b)} Training accuracy of BSD with a hint, guide, or full trace, including a full-trace variant that also weights successful rollouts.
\textbf{(c)} Avg@4 across the three benchmarks for these variants and OPSD; the base model scores 59.2.
Despite similar training-accuracy curves, richer privileged context generally yields better results, with the full trace performing best.
Appendix~\ref{app:ablation_scores} reports the per-benchmark scores.}
  \label{fig:ablation}
\end{figure*}

\textbf{Performance.}
A substantial decline in performance accompanies the response-length drift under the closed API teacher.
Figure~\ref{fig:progression}d illustrates this behavior for \texttt{Qwen3-4B-Instruct} distilled from \texttt{GPT-5.6 Luna}.
SFT is already below the base model at the beginning, and its performance falls to nearly zero by the end of training.
The additional curves in Appendix~\ref{app:dev_curves} show similar collapses or declines of SFT in other teacher--student pairs.
These results further indicate that directly imitating a stronger teacher's reasoning traces can lead to unstable training.
In contrast, ROPD and BSD maintain stable or slightly improving performance throughout training.
BSD is particularly notable because it uses the same external-teacher reasoning traces as SFT, but treats them as privileged context to guide the student's own rollouts rather than as targets to imitate.
Appendix~\ref{app:ood} further shows that SFT suffers from catastrophic forgetting on two out-of-domain benchmarks, whereas BSD does not.

\section{Further Analysis}
\label{sec:analysis}

\subsection{Ablation Study}
\label{sec:ablation}

BSD combines privileged context from an external teacher with verifier-guided learning and applies token-level guidance only to unsuccessful rollouts.
We ablate these design choices by asking three questions: (1) \emph{whether privileged context itself improves learning}, (2) \emph{how much reasoning the context should contain}, and (3) \emph{whether the verifier and the restriction to unsuccessful rollouts are needed}.
We study these questions in the shared-tokenizer setting using \texttt{Qwen3-4B} as the student and \texttt{Qwen3-32B} as the external teacher.
All variants use the same training problems and optimization setup.

\textbf{BSD-trace}, which uses the full reasoning trace as in the main experiments, serves as the reference.
For (1), we compare it with \textbf{GRPO}~\citep{shao2024deepseekmath}, which has no privileged context or self-teacher, and \textbf{RLSD}~\citep{yang2026selfdistilled}, which uses only the ground-truth answer while otherwise following BSD.
For (2), \textbf{BSD-hint} uses a three-sentence hint and \textbf{BSD-guide} uses up to eight reasoning steps without computations, both followed by the answer and derived from the full trace (obtained by using the prompts in Appendix~\ref{app:prompts}).
For (3), \textbf{BSD (w/ succ.)} additionally weights successful rollouts, while \textbf{OPSD}~\citep{zhao2026opsd} uses the self-teacher's token-level signal on every rollout without a verifier.
Figure~\ref{fig:ablation} reports training accuracy in panels (a)--(b) and Avg@4 in panel (c).

\textbf{Privileged context facilitates reinforcement learning.}
GRPO, RLSD, and BSD-trace exhibit nearly identical training-accuracy curves, with their average differences remaining below 0.2, 
although BSD-trace shows a slight advantage in the middle of training (Figure~\ref{fig:ablation}a).
Nevertheless, Avg@4 increases from 59.2 for the base model to 60.6 with GRPO, 61.7 with answer-conditioned RLSD, and 63.9 with trace-conditioned BSD.
Privileged context therefore improves how verifier feedback transfers to downstream problems.

\textbf{More complete reasoning produces stronger gains.}
Changing the privileged context from a short hint to a full trace has little effect on the training dynamics, as the three variants follow similar accuracy curves (Figure~\ref{fig:ablation}b).
However, BSD-hint scores 61.7, while BSD-guide reaches 62.5 and BSD-trace further improves to 63.9 (Figure~\ref{fig:ablation}c), indicating that downstream performance benefits from more detailed reasoning in the privileged context.

\textbf{The verifier and the restriction to unsuccessful rollouts matter.}
BSD (w/ succ.) scores 61.4, below the 63.9 of BSD-trace, so extending the weight to successful rollouts does not help (Figure~\ref{fig:ablation}c).
OPSD scores 59.7 and declines over training, from 63.3 at step 25 to 55.8 at step 150, while its responses on the benchmarks shorten from 16.8k to 12.5k tokens (Appendix~\ref{app:ablation_scores}).
The other variants keep the response length of the base model.

\subsection{What the Token-Level Guidance Captures}
\label{sec:anatomy}

BSD uses the token-level guidance $\delta_t$ to distribute verifier feedback across the tokens of an unsuccessful rollout.
To understand this signal, we examine which tokens it emphasizes, whether its strength is associated with BSD's gains, and how it affects the student's behavior.
Table~\ref{tab:guidance} reports the quantitative results averaged across teachers, with the measurement setup described in Appendix~\ref{app:exp_details}, while Figure~\ref{fig:case} illustrates a concrete example.

\begin{table}[t]
\setlength{\belowcaptionskip}{-0.2cm}
\centering
\renewcommand{\arraystretch}{0.88}
\small
\begin{tabular}{lccc}
\toprule
Student & $|\bar\delta|$ & $\rho_H$ & Gain \\
\midrule
\texttt{Qwen3-0.6B} & 0.072 & 0.87 & $+2.2$ \\
\texttt{Qwen3-1.7B} & 0.115 & 0.90 & $+3.5$ \\
\texttt{Qwen3-4B} & 0.114 & 0.90 & $+3.9$ \\
\texttt{Qwen3-4B-Instruct} & 0.037 & 0.83 & $+1.5$ \\
\bottomrule
\end{tabular}
\caption{Token-level guidance before training, measured on unsuccessful rollouts.
$|\bar\delta|$ is the absolute value of the mean of $\delta_t$ over all tokens, 
$\rho_H$ is the Spearman correlation between $|\delta_t|$ and the student's token entropy, and Gain is the mean improvement of BSD over the base model shown in Table~\ref{tab:main}.}
\label{tab:guidance}
\end{table}

\textbf{Guidance targets uncertain tokens and tracks performance gains.}
The correlation $\rho_H$ between per-token guidance magnitude and the student's token entropy on unsuccessful rollouts ranges from $0.83$ to $0.90$, indicating that the strongest guidance concentrates on tokens where the student is most uncertain.
Its magnitude also follows the gains: \texttt{Qwen3-4B-Instruct} has both the weakest guidance and the smallest gain, whereas \texttt{Qwen3-1.7B} and \texttt{Qwen3-4B} receive stronger guidance and achieve larger gains.
Thus, BSD achieves larger gains when the privileged context more strongly changes the student's predictions, particularly at uncertain tokens.

\textbf{Case study: BSD encourages self-correction.}
Figure~\ref{fig:case} shows an AIME24 problem where the base \texttt{Qwen3-4B} finds the correct integer $N=5694$ but misreads its quotient and remainder when it is divided by $1000$.
Given the same prefix containing this mistake, the base model continues without revisiting the division, whereas the student trained with BSD re-checks the step, identifies the error, and reaches the correct answer.
Across eight continuations from this prefix, the BSD student solves the problem six times, compared with four for the base model.
This example illustrates how BSD can help the student revisit and correct an uncertain intermediate step rather than persist with an initially plausible but incorrect rollout.\footnote{Appendix~\ref{app:wording} further compares the wording of responses from the base model and the BSD-trained student, and Appendix~\ref{app:ablation_scores} shows an example of the token-level guidance under the trace and under the answer.}

\begin{figure}[t]
  \centering
  \setlength{\belowcaptionskip}{-0.3cm}
  \footnotesize
  \setlength{\fboxsep}{4pt}
  \fcolorbox{black!35}{white}{\parbox{0.94\columnwidth}{%
    \textbf{Problem} (AIME24). $N$ is the greatest four-digit integer such
    that changing any one of its digits to $1$ gives a multiple of $7$.
    $Q$ and $R$ are the quotient and remainder when $N$ is divided by
    $1000$. Find $Q+R$. \hfill Answer: $699$}}\\[3pt]
  \fcolorbox{black!35}{black!7}{\parbox{0.94\columnwidth}{%
    \textbf{Shared prefix: a rollout of the base model, first 90\%.}\\{}
    [\ldots] All divisible by 7. So this works. [\ldots] Therefore, it seems
    that 5694 is the only possible number, and hence the answer is
    \textcolor{wrongred}{\textbf{Q=569, R=4, so Q+R=573}}. [\ldots]}}\\[3pt]
  \fcolorbox{black!35}{white}{\parbox{0.94\columnwidth}{%
    \textbf{The base model continues.}\\{}
    [\ldots] It seems that no higher number works. Therefore, 5694 is indeed
    the maximum. [\ldots] Therefore, the answer is Q=569, R=4, so
    Q + R = 573. [\ldots] $Q = \lfloor 5694/1000 \rfloor = 569$,
    $R = 5694 \bmod 1000 = 694$. Hence, $Q + R = 569 + 694 = 1263$.
\\{}
    \textbf{Final answer:} $1263$ \textcolor{wrongred}{\ding{55}}}}\\[3pt]
  \fcolorbox{bsdblue}{white}{\parbox{0.94\columnwidth}{%
    \textbf{\textcolor{bsdblue}{The student trained with BSD continues.}}\\{}
    [\ldots] Thus, the final answer is Q=569, R=4, so 569 + 4 = 573.
    \textcolor{bsdblue}{\textbf{But wait, let me double-check}} the
    calculation of N divided by 1000. N=5694. When divided by 1000, the
    quotient Q is 5 (since 5694 = 5*1000 + 694), and the remainder R is
    694. [\ldots] \textcolor{bsdblue}{\textbf{Wait, this is a critical
    mistake!}} [\ldots] So Q=5, R=694, so Q+R=5 + 694=699.
\\{}
    \textbf{Final answer:} $699$ \textcolor{bsdblue}{\ding{51}}}}
      \caption{Case study on an AIME 2024 problem.
From the same erroneous prefix, the base \texttt{Qwen3-4B} continues incorrectly, whereas the BSD-trained student (\texttt{Qwen3-32B} teacher) checks the quotient and remainder and reaches the correct answer.
[\ldots] marks omissions.}
  \label{fig:case}
\end{figure}

\section{Related Work}
\label{sec:related}

\textbf{Knowledge distillation.}
Knowledge distillation transfers supervision from a stronger teacher to a smaller student through output distributions or generated sequences \citep{hinton2015distilling,kim2016sequence}. 
For reasoning models, this idea is commonly instantiated by training on teacher-generated reasoning traces, although large teacher--student gaps can make direct imitation difficult \citep{shridhar-etal-2023-distilling,hsieh-etal-2023-distilling,zhang-etal-2023-lifting,huang2026finetunereasoningmodelteacherstudent,lyu2026studentcentereddistillationnarrowsagentic}.
On-policy distillation instead learns from student-generated rollouts under token-level teacher guidance \citep{gu2024minillm,agarwal2024gkd,jin2026entropyopd,yang2026genopd}.
Recent variants extend OPD to heterogeneous tokenizers \citep{minixhofer2025universal,boizard2025crosstokenizerdistillationuniversallogit,niu2026breaking,sun2026simctrecoveringlostsupervision} and black-box teachers \citep{ye2026blackboxonpolicydistillationlarge,xiong2026ovdonpolicyverbaldistillation,fang2026ropd,zhou2026omniopd}, but typically require setting-specific supervision.
In contrast, BSD offers an effective and unified framework for all these settings.

\textbf{Privileged information and self-distillation.}
Learning with privileged information provides a general framework for exploiting supervision that is available only during training, with generalized distillation connecting this idea to teacher--student learning \citep{vapnik2009lupi,lopezpaz2016unifying}.
In language models, context distillation instantiates this principle by internalizing information from instructions, demonstrations, or other contextual signals into a model that later operates without them \citep{snell2022learningdistillingcontext,shin2025generativepromptinternalization,duan2024incontextlearningdistillationefficient,ye2026onpolicycontextdistillationlanguage}.
Recent self-distillation methods further use the model itself as a privileged teacher by conditioning it on answers, solutions, feedback, or successful trajectories \citep{zhao2026opsd,penaloza2026privilegedinformationdistillationlanguage,huebotter2026sdpo,ye2026onpolicycontextdistillationlanguage,ding2026hdpo,liu2026crosslingualonpolicyselfdistillationmultilingual,yang2026selfdistilled,pan2026rlcsd,li2026srpo,kim2026rlrt}.
BSD follows this privileged paradigm, but uses reasoning from an \emph{external teacher} as the privileged context.

\section{Conclusion}
\label{sec:conclusion}

We introduced BSD, a unified framework for distilling from any teacher that provides reasoning traces.
BSD uses the external teacher's reasoning trace as privileged context for a self-teacher, producing compatible token-level guidance in the student's own token space without teacher probabilities or vocabulary mappings.
Across twelve teacher--student pairs spanning shared-tokenizer, cross-tokenizer, and closed API settings, BSD improves every base model and achieves the best or tied average performance among the distillation methods in eleven pairs.
Our analyses further show that BSD supports stable training, benefits from more complete reasoning, and focuses its guidance on uncertain tokens, establishing reasoning traces as a common and effective interface for on-policy distillation.

\section*{Limitations}

We evaluate BSD with shared-tokenizer, cross-tokenizer, and closed API
teachers across twelve teacher--student pairs and analyze its token-level
guidance. However, we acknowledge three main limitations.

First, due to computational constraints, our experiments are limited to
students of up to 4B parameters from one model family and to English
competition mathematics. Future work could examine whether the gains of
BSD hold for larger students, other model families, and other domains.

Second, BSD relies on a verifier to set the direction of each update, so
it applies only to tasks whose answers can be checked automatically and
not to open-ended ones. Whether the agreement between the student and the
external teacher can serve as a verifier in such tasks remains to be
explored.

Finally, our analysis relates the token-level guidance to the gains of
BSD only through correlations across students. It characterizes where the
guidance concentrates but does not establish which tokens it helps, or
why, which we leave for future work.

\section*{Ethics Statement}
This work fine-tunes publicly released language models on publicly available
mathematical reasoning datasets and evaluates them with automatic verifiers.
It involves no human subjects, no personally identifiable information, and no
newly collected data. 
The method transfers reasoning capability that the
teacher models already expose through their public interfaces, so we do not
identify risks beyond those general to improving open-weight reasoning
models.

The closed API teacher is queried through its official API solely to
obtain responses for non-commercial research. The students trained with
these responses serve only to study distillation methods: we do not
deploy them, offer them as a service, or use them to develop models that
compete with the API provider. We release only code, and release no
model weights and no datasets, including the teacher's reasoning traces.

\textbf{Use of AI Assistants.}
The authors used AI assistants for code development, the organization of
experiments and analyses, figure preparation, and manuscript drafting and
language editing. All reported experimental results and scientific
interpretations were reviewed by the authors, who take responsibility for
the final paper.

\section*{Acknowledgments}
This project was generously funded by the German Federal Justice Ministry
as part of the Digitalisierungsinitiative des Bundes f{\"u}r die Justiz
under the GSJ Project (``Generatives Sprachmodell der Justiz''), a
collaboration between the justice ministries of Northrhine-Westphalia and
Bavaria, TU Munich, and the University of Cologne.
The authors gratefully acknowledge the scientific support and resources of
the AI service infrastructure LRZ AI Systems provided by the Leibniz
Supercomputing Centre (LRZ) of the Bavarian Academy of Sciences and
Humanities (BAdW), funded by the Bayerisches Staatsministerium f{\"u}r
Wissenschaft und Kunst (StMWK), primarily through the BayernKI initiative.
This research was also supported by the Munich Center for Machine Learning 
(MCML) and German Research Foundation (DFG, grant SCHU 2246/14-1).

\bibliography{references}

\begin{thebibliography}{67}
\providecommand{\natexlab}[1]{#1}

\bibitem[{Agarwal et~al.(2024)Agarwal, Vieillard, Zhou, Stanczyk, Garea, Geist,
  and Bachem}]{agarwal2024gkd}
Rishabh Agarwal, Nino Vieillard, Yongchao Zhou, Piotr Stanczyk, Sabela~Ramos
  Garea, Matthieu Geist, and Olivier Bachem. 2024.
\newblock \href {https://openreview.net/forum?id=3zKtaqxLhW} {On-policy
  distillation of language models: Learning from self-generated mistakes}.
\newblock In \emph{The Twelfth International Conference on Learning
  Representations, {ICLR} 2024, Vienna, Austria, May 7-11, 2024}.
  OpenReview.net.

\bibitem[{Ahn et~al.(2024)Ahn, Verma, Lou, Liu, Zhang, and
  Yin}]{ahn-etal-2024-large}
Janice Ahn, Rishu Verma, Renze Lou, Di~Liu, Rui Zhang, and Wenpeng Yin. 2024.
\newblock \href {https://doi.org/10.18653/v1/2024.eacl-srw.17} {Large language
  models for mathematical reasoning: Progresses and challenges}.
\newblock In \emph{Proceedings of the 18th Conference of the European Chapter
  of the Association for Computational Linguistics: Student Research Workshop},
  pages 225--237, St. Julian{'}s, Malta. Association for Computational
  Linguistics.

\bibitem[{Boizard et~al.(2025)Boizard, Haddad, Hudelot, and
  Colombo}]{boizard2025crosstokenizerdistillationuniversallogit}
Nicolas Boizard, Kevin~El Haddad, C{\'{e}}line Hudelot, and Pierre Colombo.
  2025.
\newblock \href {https://openreview.net/forum?id=bwRxXiGO9A} {Towards
  cross-tokenizer distillation: the universal logit distillation loss for
  {LLMs}}.
\newblock \emph{Trans. Mach. Learn. Res.}, 2025.

\bibitem[{Chung et~al.(2024)Chung, Hou, Longpre, Zoph, Tay, Fedus, Li, Wang,
  Dehghani, Brahma, Webson, Gu, Dai, Suzgun, Chen, Chowdhery, Castro{-}Ros,
  Pellat, Robinson, Valter, Narang, Mishra, Yu, Zhao, Huang, Dai, Yu, Petrov,
  Chi, Dean, Devlin, Roberts, Zhou, Le, and Wei}]{Chung2024sft}
Hyung~Won Chung, Le~Hou, Shayne Longpre, Barret Zoph, Yi~Tay, William Fedus,
  Yunxuan Li, Xuezhi Wang, Mostafa Dehghani, Siddhartha Brahma, Albert Webson,
  Shixiang~Shane Gu, Zhuyun Dai, Mirac Suzgun, Xinyun Chen, Aakanksha
  Chowdhery, Alex Castro{-}Ros, Marie Pellat, Kevin Robinson, and 16 others.
  2024.
\newblock \href {https://jmlr.org/papers/v25/23-0870.html} {Scaling
  instruction-finetuned language models}.
\newblock \emph{J. Mach. Learn. Res.}, 25:70:1--70:53.

\bibitem[{Dekoninck et~al.(2026)Dekoninck, Jovanović, Gehrunger,
  Rögnvaldsson, Petrov, Sun, and Vechev}]{dekoninck2026matharena}
Jasper Dekoninck, Nikola Jovanović, Tim Gehrunger, Kári Rögnvaldsson, Ivo
  Petrov, Chenhao Sun, and Martin Vechev. 2026.
\newblock \href {https://arxiv.org/abs/2605.00674} {Beyond benchmarks:
  {MathArena} as an evaluation platform for mathematics with {LLMs}}.
\newblock \emph{Preprint}, arXiv:2605.00674.

\bibitem[{Ding(2026)}]{ding2026hdpo}
Ken Ding. 2026.
\newblock \href {https://arxiv.org/abs/2603.23871} {{HDPO}: Hybrid distillation
  policy optimization via privileged self-distillation}.
\newblock \emph{Preprint}, arXiv:2603.23871.

\bibitem[{Duan et~al.(2024)Duan, Li, Zhai, and
  Yao}]{duan2024incontextlearningdistillationefficient}
Yifei Duan, Liu Li, Zirui Zhai, and Jinxia Yao. 2024.
\newblock \href {https://arxiv.org/abs/2412.13243} {In-context learning
  distillation for efficient few-shot fine-tuning}.
\newblock \emph{Preprint}, arXiv:2412.13243.

\bibitem[{Fan et~al.(2026)Fan, Ni, Merane, Tian, Hermstr{\"{u}}wer, Huang,
  Akhtar, Salimbeni, Geering, Dreyer, Brunner, Leippold, Sachan, Stremitzer,
  Engel, Ash, and Niklaus}]{fan2026lexam}
Yu~Fan, Jingwei Ni, Jakob Merane, Yang Tian, Yoan Hermstr{\"{u}}wer, Yinya
  Huang, Mubashara Akhtar, Etienne Salimbeni, Florian Geering, Oliver Dreyer,
  Daniel Brunner, Markus Leippold, Mrinmaya Sachan, Alexander Stremitzer,
  Christoph Engel, Elliott Ash, and Joel Niklaus. 2026.
\newblock \href
  {https://proceedings.iclr.cc/paper\_files/paper/2026/hash/47cbdfce2fc71cd473bd7d409cd1bf4b-Abstract-Conference.html}
  {{LEXam}: Benchmarking legal reasoning on 340 law exams}.
\newblock In \emph{The Fourteenth International Conference on Learning
  Representations, {ICLR} 2026, Rio de Janeiro, Brazil, April 23-27, 2026}.
  proceedings.iclr.cc.

\bibitem[{Fang et~al.(2026)Fang, Hong, Zheng, Song, Li, Jiang, Zhang, Guo,
  Wang, and Chua}]{fang2026ropd}
Junfeng Fang, Zhepei Hong, Mao Zheng, Mingyang Song, Gengsheng Li, Houcheng
  Jiang, Dan Zhang, Haiyun Guo, Xiang Wang, and Tat-Seng Chua. 2026.
\newblock \href {https://arxiv.org/abs/2605.07396} {Rubric-based on-policy
  distillation}.
\newblock \emph{Preprint}, arXiv:2605.07396.

\bibitem[{Gu et~al.(2024)Gu, Dong, Wei, and Huang}]{gu2024minillm}
Yuxian Gu, Li~Dong, Furu Wei, and Minlie Huang. 2024.
\newblock \href {https://openreview.net/forum?id=5h0qf7IBZZ} {{MiniLLM}:
  Knowledge distillation of large language models}.
\newblock In \emph{The Twelfth International Conference on Learning
  Representations, {ICLR} 2024, Vienna, Austria, May 7-11, 2024}.
  OpenReview.net.

\bibitem[{Guha et~al.(2026)Guha, Marten, Keh, Raoof, Smyrnis, Bansal,
  Nezhurina, Mercat, Vu, Sprague, Suvarna, Feuer, Chen, Khan, Frankel, Grover,
  Choi, Muennighoff, Su, Zhao, Yang, Pimpalgaonkar, sharma, Ji, Deng, Pratt,
  Ramanujan, Saad-Falcon, Acharya, Li, Dave, Albalak, Arora, Wulfe, Hegde,
  Durrett, Oh, Bansal, Gabriel, Grover, Chang, Shankar, Gokaslan, Merrill,
  Hashimoto, Choi, Jitsev, Heckel, Sathiamoorthy, Dimakis, and
  Schmidt}]{guha2025openthoughts}
Etash Guha, Ryan Marten, Sedrick Keh, Negin Raoof, Georgios Smyrnis, Hritik
  Bansal, Marianna Nezhurina, Jean Mercat, Trung Vu, Zayne Sprague, Ashima
  Suvarna, Benjamin Feuer, Leon~Liangyu Chen, Zaid Khan, Eric Frankel, Sachin
  Grover, Caroline Choi, Niklas Muennighoff, Shiye Su, and 32 others. 2026.
\newblock \href
  {https://proceedings.iclr.cc/paper_files/paper/2026/file/b010241b9f1cdfc7d4c392db899cef86-Paper-Conference.pdf}
  {{OpenThoughts}: Data recipes for reasoning models}.
\newblock In \emph{International Conference on Learning Representations},
  volume 2026, pages 108059--108130.

\bibitem[{Guo et~al.(2025)Guo, Yang, Zhang, Song, Wang, Zhu, Xu, Zhang, Ma, Bi,
  Zhang, Yu, Wu, Wu, Gou, Shao, Li, Gao, Liu, Xue, Wang, Wu, Feng, Lu, Zhao,
  Deng, Ruan, Dai, Chen, Ji, Li, Lin, Dai, Luo, Hao, Chen, Li, Zhang, Xu, Ding,
  Gao, Qu, Li, Guo, Li, Chen, Yuan, Tu, Qiu, Li, Cai, Ni, Liang, Chen, Dong,
  Hu, You, Gao, Guan, Huang, Yu, Wang, Zhang, Zhao, Wang, Zhang, Xu, Xia,
  Zhang, Zhang, Tang, Zhou, Li, Wang, Li, Tian, Huang, Zhang, Wang, Chen, Du,
  Ge, Zhang, Pan, Wang, Chen, Jin, Chen, Lu, Zhou, Chen, Ye, Wang, Yu, Zhou,
  Pan, Li, Zhou, Wu, Yun, Pei, Sun, Wang, Zeng, Liu, Liang, Gao, Yu, Zhang,
  Xiao, An, Liu, Wang, Chen, Nie, Cheng, Liu, Xie, Liu, Yang, Li, Su, Lin, Li,
  Jin, Shen, Chen, Sun, Wang, Song, Zhou, Wang, Shan, Li, Wang, Wei, Zhang, Xu,
  Li, Zhao, Sun, Wang, Yu, Zhang, Shi, Xiong, He, Piao, Wang, Tan, Ma, Liu,
  Guo, Ou, Wang, Gong, Zou, He, Xiong, Luo, You, Liu, Zhou, Zhu, Huang, Li,
  Zheng, Zhu, Ma, Tang, Zha, Yan, Ren, Ren, Sha, Fu, Xu, Xie, Zhang, Hao, Ma,
  Yan, Wu, Gu, Zhu, Liu, Li, Xie, Song, Pan, Huang, Xu, Zhang, and
  Zhang}]{deepseekr1}
Daya Guo, Dejian Yang, Haowei Zhang, Junxiao Song, Peiyi Wang, Qihao Zhu,
  Runxin Xu, Ruoyu Zhang, Shirong Ma, Xiao Bi, Xiaokang Zhang, Xingkai Yu,
  Yu~Wu, Z.~F. Wu, Zhibin Gou, Zhihong Shao, Zhuoshu Li, Ziyi Gao, Aixin Liu,
  and 175 others. 2025.
\newblock \href {https://doi.org/10.1038/s41586-025-09422-z} {{DeepSeek-R1}
  incentivizes reasoning in {LLMs} through reinforcement learning}.
\newblock \emph{Nature}, 645(8081):633–638.

\bibitem[{Hinton et~al.(2015)Hinton, Vinyals, and Dean}]{hinton2015distilling}
Geoffrey Hinton, Oriol Vinyals, and Jeff Dean. 2015.
\newblock \href {https://arxiv.org/abs/1503.02531} {Distilling the knowledge in
  a neural network}.
\newblock \emph{Preprint}, arXiv:1503.02531.

\bibitem[{Hsieh et~al.(2023)Hsieh, Li, Yeh, Nakhost, Fujii, Ratner, Krishna,
  Lee, and Pfister}]{hsieh-etal-2023-distilling}
Cheng-Yu Hsieh, Chun-Liang Li, Chih-kuan Yeh, Hootan Nakhost, Yasuhisa Fujii,
  Alex Ratner, Ranjay Krishna, Chen-Yu Lee, and Tomas Pfister. 2023.
\newblock \href {https://doi.org/10.18653/v1/2023.findings-acl.507} {Distilling
  step-by-step! outperforming larger language models with less training data
  and smaller model sizes}.
\newblock In \emph{Findings of the Association for Computational Linguistics:
  ACL 2023}, pages 8003--8017, Toronto, Canada. Association for Computational
  Linguistics.

\bibitem[{Hu et~al.(2022)Hu, Shen, Wallis, Allen{-}Zhu, Li, Wang, Wang, and
  Chen}]{hu2022lora}
Edward~J. Hu, Yelong Shen, Phillip Wallis, Zeyuan Allen{-}Zhu, Yuanzhi Li,
  Shean Wang, Lu~Wang, and Weizhu Chen. 2022.
\newblock \href {https://openreview.net/forum?id=nZeVKeeFYf9} {{LoRA}: Low-rank
  adaptation of large language models}.
\newblock In \emph{The Tenth International Conference on Learning
  Representations, {ICLR} 2022, Virtual Event, April 25-29, 2022}.
  OpenReview.net.

\bibitem[{Huang et~al.(2026)Huang, Yang, Huang, Hao, Ge, Li, Du, Chen, and
  Guo}]{huang2026finetunereasoningmodelteacherstudent}
Zixian Huang, Kaichen Yang, Xu~Huang, Feiyang Hao, Qiming Ge, Bowen Li, He~Du,
  Kai Chen, and Qipeng Guo. 2026.
\newblock \href {https://arxiv.org/abs/2604.14164} {How to fine-tune a
  reasoning model? a teacher-student cooperation framework to synthesize
  student-consistent {SFT} data}.
\newblock \emph{Preprint}, arXiv:2604.14164.

\bibitem[{Hübotter et~al.(2026)Hübotter, Lübeck, Behric, Baumann, Bagatella,
  Marta, Hakimi, Shenfeld, Buening, Guestrin, and Krause}]{huebotter2026sdpo}
Jonas Hübotter, Frederike Lübeck, Lejs Behric, Anton Baumann, Marco
  Bagatella, Daniel Marta, Ido Hakimi, Idan Shenfeld, Thomas~Kleine Buening,
  Carlos Guestrin, and Andreas Krause. 2026.
\newblock \href {https://arxiv.org/abs/2601.20802} {Reinforcement learning via
  self-distillation}.
\newblock \emph{Preprint}, arXiv:2601.20802.

\bibitem[{Jiang et~al.(2023)Jiang, Chan, Chen, and Wang}]{jiang-etal-2023-lion}
Yuxin Jiang, Chunkit Chan, Mingyang Chen, and Wei Wang. 2023.
\newblock \href {https://doi.org/10.18653/v1/2023.emnlp-main.189} {{L}ion:
  Adversarial distillation of proprietary large language models}.
\newblock In \emph{Proceedings of the 2023 Conference on Empirical Methods in
  Natural Language Processing}, pages 3134--3154, Singapore. Association for
  Computational Linguistics.

\bibitem[{Jin et~al.(2026)Jin, Min, Yang, Wei, Zhou, Kadhe, Baracaldo, and
  Lee}]{jin2026entropyopd}
Woogyeol Jin, Taywon Min, Yongjin Yang, Dennis Wei, Yi~Zhou, Swanand~Ravindra
  Kadhe, Nathalie Baracaldo, and Kimin Lee. 2026.
\newblock \href {https://openreview.net/forum?id=J5i09faOOf} {Entropy-aware
  on-policy distillation of language models}.
\newblock In \emph{Forty-third International Conference on Machine Learning}.

\bibitem[{Kim et~al.(2026)Kim, Jeon, Li, and Yang}]{kim2026rlrt}
Jeonghye Kim, Jiwon Jeon, Dongsheng Li, and Yuqing Yang. 2026.
\newblock \href {https://arxiv.org/abs/2605.10781} {Rebellious student:
  Reversing teacher signals for reasoning exploration with self-distilled
  {RLVR}}.
\newblock \emph{Preprint}, arXiv:2605.10781.

\bibitem[{Kim and Rush(2016)}]{kim2016sequence}
Yoon Kim and Alexander~M. Rush. 2016.
\newblock \href {https://doi.org/10.18653/v1/D16-1139} {Sequence-level
  knowledge distillation}.
\newblock In \emph{Proceedings of the 2016 Conference on Empirical Methods in
  Natural Language Processing}, pages 1317--1327, Austin, Texas. Association
  for Computational Linguistics.

\bibitem[{Ko et~al.(2024)Ko, Kim, Chen, and Yun}]{ko2024distillm}
Jongwoo Ko, Sungnyun Kim, Tianyi Chen, and Se{-}Young Yun. 2024.
\newblock \href {https://proceedings.mlr.press/v235/ko24c.html} {{DistiLLM}:
  Towards streamlined distillation for large language models}.
\newblock In \emph{Forty-first International Conference on Machine Learning,
  {ICML} 2024, Vienna, Austria, July 21-27, 2024}, volume 235 of
  \emph{Proceedings of Machine Learning Research}, pages 24872--24895. {PMLR} /
  OpenReview.net.

\bibitem[{Kwon et~al.(2023)Kwon, Li, Zhuang, Sheng, Zheng, Yu, Gonzalez, Zhang,
  and Stoica}]{kwon2023vllm}
Woosuk Kwon, Zhuohan Li, Siyuan Zhuang, Ying Sheng, Lianmin Zheng, Cody~Hao Yu,
  Joseph Gonzalez, Hao Zhang, and Ion Stoica. 2023.
\newblock \href {https://doi.org/10.1145/3600006.3613165} {Efficient memory
  management for large language model serving with {PagedAttention}}.
\newblock In \emph{Proceedings of the 29th Symposium on Operating Systems
  Principles, {SOSP} 2023, Koblenz, Germany, October 23-26, 2023}, pages
  611--626. {ACM}.

\bibitem[{Li et~al.(2026{\natexlab{a}})Li, Yang, Fang, Song, Zheng, Guo, Zhang,
  Wang, and Chua}]{li2026srpo}
Gengsheng Li, Tianyu Yang, Junfeng Fang, Mingyang Song, Mao Zheng, Haiyun Guo,
  Dan Zhang, Jinqiao Wang, and Tat-Seng Chua. 2026{\natexlab{a}}.
\newblock \href {https://openreview.net/forum?id=P2OuWwZspP} {Unifying
  group-relative and self-distillation policy optimization via sample routing}.
\newblock In \emph{Third Conference on Language Modeling}.

\bibitem[{Li et~al.(2026{\natexlab{b}})Li, Zuo, He, Zhang, Xiao, Qian, Yu, ang
  Gao, Yang, Liu, and Ding}]{li2026rethinkingopd}
Yaxuan Li, Yuxin Zuo, Bingxiang He, Jinqian Zhang, Chaojun Xiao, Cheng Qian,
  Tianyu Yu, Huan ang Gao, Wenkai Yang, Zhiyuan Liu, and Ning Ding.
  2026{\natexlab{b}}.
\newblock \href {https://arxiv.org/abs/2604.13016} {Rethinking on-policy
  distillation of large language models: Phenomenology, mechanism, and recipe}.
\newblock \emph{Preprint}, arXiv:2604.13016.

\bibitem[{Liu et~al.(2026)Liu, Zhao, Hedderich, and
  Schütze}]{liu2026crosslingualonpolicyselfdistillationmultilingual}
Yihong Liu, Raoyuan Zhao, Michael~A. Hedderich, and Hinrich Schütze. 2026.
\newblock \href {https://arxiv.org/abs/2605.09548} {Crosslingual on-policy
  self-distillation for multilingual reasoning}.
\newblock \emph{Preprint}, arXiv:2605.09548.

\bibitem[{Lopez{-}Paz et~al.(2016)Lopez{-}Paz, Bottou, Sch{\"{o}}lkopf, and
  Vapnik}]{lopezpaz2016unifying}
David Lopez{-}Paz, L{\'{e}}on Bottou, Bernhard Sch{\"{o}}lkopf, and Vladimir
  Vapnik. 2016.
\newblock \href {https://doi.org/10.48550/ARXIV.1511.03643} {Unifying
  distillation and privileged information}.
\newblock In \emph{4th International Conference on Learning Representations,
  {ICLR} 2016, San Juan, Puerto Rico, May 2-4, 2016, Conference Track
  Proceedings}.

\bibitem[{Lu and Lab(2025)}]{lu2025onpolicyblog}
Kevin Lu and Thinking~Machines Lab. 2025.
\newblock \href {https://doi.org/10.64434/tml.20251026} {On-policy
  distillation}.
\newblock \emph{Thinking Machines Lab: Connectionism}.

\bibitem[{Lyu et~al.(2026)Lyu, Wang, Huang, and
  Xu}]{lyu2026studentcentereddistillationnarrowsagentic}
Yuanjie Lyu, Chengyu Wang, Jun Huang, and Tong Xu. 2026.
\newblock \href {https://openreview.net/forum?id=KaTYG9LGJv} {Student-centered
  distillation narrows the agentic gap between small and large {LLM}s}.
\newblock In \emph{Forty-third International Conference on Machine Learning}.

\bibitem[{{Mathematical Association of America}(2024)}]{aime24}
{Mathematical Association of America}. 2024.
\newblock \href {https://huggingface.co/datasets/HuggingFaceH4/aime_2024}
  {{American Invitational Mathematics Examination} ({AIME}) 2024}.
\newblock Problems from the HuggingFaceH4/aime\_2024 dataset.

\bibitem[{{Mathematical Association of America}(2025)}]{aime25}
{Mathematical Association of America}. 2025.
\newblock \href {https://huggingface.co/datasets/yentinglin/aime_2025}
  {{American Invitational Mathematics Examination} ({AIME}) 2025}.
\newblock Problems from the yentinglin/aime\_2025 dataset.

\bibitem[{Minixhofer et~al.(2025)Minixhofer, Vulic, and
  Ponti}]{minixhofer2025universal}
Benjamin Minixhofer, Ivan Vulic, and Edoardo~Maria Ponti. 2025.
\newblock \href
  {http://papers.nips.cc/paper\_files/paper/2025/hash/720f9f5dc751eb56952ae4fee2398f73-Abstract-Conference.html}
  {Universal cross-tokenizer distillation via approximate likelihood matching}.
\newblock In \emph{Advances in Neural Information Processing Systems 38: Annual
  Conference on Neural Information Processing Systems 2025, NeurIPS 2025, San
  Diego, CA, USA, December 2-7, 2025 / Mexico City, Mexico, November 30 -
  December 5, 2025}.

\bibitem[{Muennighoff et~al.(2025)Muennighoff, Yang, Shi, Li, Fei-Fei,
  Hajishirzi, Zettlemoyer, Liang, Cand{\`e}s, and
  Hashimoto}]{muennighoff-etal-2025-s1}
Niklas Muennighoff, Zitong Yang, Weijia Shi, Xiang~Lisa Li, Li~Fei-Fei,
  Hannaneh Hajishirzi, Luke Zettlemoyer, Percy Liang, Emmanuel Cand{\`e}s, and
  Tatsunori Hashimoto. 2025.
\newblock \href {https://doi.org/10.18653/v1/2025.emnlp-main.1025} {s1: Simple
  test-time scaling}.
\newblock In \emph{Proceedings of the 2025 Conference on Empirical Methods in
  Natural Language Processing}, pages 20275--20321, Suzhou, China. Association
  for Computational Linguistics.

\bibitem[{Niu et~al.(2026)Niu, Xiao, Liu, Wang, Gong, Wang, and
  Li}]{niu2026breaking}
Yifan Niu, Han Xiao, Dongyi Liu, Zelong Wang, Dihong Gong, Yasheng Wang, and
  Jia Li. 2026.
\newblock \href {https://arxiv.org/abs/2606.09456} {Breaking the tokenizer
  barrier: On-policy distillation across model families}.
\newblock \emph{Preprint}, arXiv:2606.09456.

\bibitem[{{NVIDIA} et~al.(2025){NVIDIA}, Blakeman, Grattafiori, Basant, Gupta,
  Khattar, Renduchintala, Vavre, Shukla, Bercovich, Ficek, Shaposhnikov,
  Kondratenko, Bukharin, Milesi, Taghibakhshi, Liu, Barton, Mahabaleshwarkar,
  Klein, Zuker, Geifman, Shen, Bhiwandiwalla, Tao, Guan, Mandarwal, Mehta,
  Aithal, Poojary, Ahamed, Thekkumpate, Dattagupta, Zhu, Sadeghi, Simkin,
  Lanir, Schifferer, Nushi, Kartal, Rouhani, Ginsburg, Norick, Soubasis,
  Kisacanin, Yu, Catanzaro, del Mundo, Hwang, Wang, Hsieh, Zhang, Yu, Mungekar,
  Patel, Alexiuk, Parisien, Neale, Mosk-Aoyama, Su, Corneil, Afrimi, Rohrer,
  Serebrenik, Gitman, Levy, Stosic, Mosallanezhad, Narayanan, Nathawani,
  Rekesh, Yared, Kakwani, Ahn, Riach, Stosic, Minasyan, Lin, Long, Long, Lantz,
  Evans, Ning, Chung, Harper, Tramel, Galinkin, Pounds, Briones, Bakhturina,
  Ladhak, Wang, Jia, Soares, Chen, Galko, Siino, Agam, Ajjanagadde, Bhatt,
  Prasad, Armstrong, Shen, Batmaz, Nalbandyan, Qian, Sharma, Ross, Ngo, Sahota,
  Wang, Soni, Upadhyay, Mao, Nguyen, Nguyen, Cunningham, Shahaf, Gitman,
  Loshchilov, Moshkov, Putterman, Kautz, Scowcroft, Casper, Mitra, Glick, Chen,
  Oliver, Zhang, Zeng, Lou, Zhang, Huang, Conway, Guman, Kamalu, Greco, Cohen,
  Jennings, Daw, Vialard, Yi, Parmar, Xu, Zhu, Briski, Cheung, Luna, Santhanam,
  Shih, Kong, Bhardwaj, Puvvada, Pawelec, Anik, McAfee, Sleiman, Derczynski,
  Ding, Liebenwein, Vega, Grover, Segbroeck, de~Melo, Sreedhar, Kilaru,
  Ashkenazi, Romeijn, Cai, Kliegl, Moosaei, Novikov, Samadi, Corpuz, Wang,
  Price, Boone, Evans, Martinez, Chrzanowski, Shoeybi, Patwary, Mulepati,
  Hereth, Assaf, Habibi, Zmora, Haber, Sessions, Bhatia, Jukar, Pope, Ludwig,
  Tajbakhsh, Juluru, Hrinchuk, Kuchaiev, Delalleau, Olabiyi, Argov, Xie,
  Chadha, Shamis, Molchanov, Morkisz, Dykas, Jin, Xu, Januszewski, Thombre,
  Varshney, Gundecha, Miao, Mahabadi, El-Yaniv, Zilberstein, Shafipour, Harang,
  Izzo, Shahbazyan, Garg, Borkar, Gala, Islam, Waleffe, Watve, Koren, Zhang,
  Hewett, Prenger, Timbrook, Mahdavi, Modi, Kriman, Kariyappa, Satheesh, Kaji,
  Pasumarthi, Narentharen, Narenthiran, Bak, Kashirsky, Poulos, Mor, Ramasamy,
  Acharya, Ghosh, Sreenivas, Thomas, Fan, Gopal, Prabhumoye, Pachori,
  Toshniwal, Ding, Singh, Sun, Ithape, Majumdar, Singhal, Alborghetti, Ge,
  Devare, Barua, Panguluri, Gupta, Priyadarshi, Akter, Bui, Ene, Kong, Do,
  Blankevoort, Balough, Asida, Natan, Konuk, Vashishth, Karpas, De, Noorozi,
  Noroozi, Srinivasan, Elango, Korthikanti, Kurin, Lavrukhin, Jiang, Ahmad, Du,
  Ping, Zhou, Jennings, Zhang, Prazuch, Ren, Karnati, Choi, Meyer, Wu, Zhang,
  Lin, Geifman, Fu, Subara, Suhara, Gao, Moshe, Dong, Liu, Chen, and
  Yan}]{nemotron3}
{NVIDIA}, Aaron Blakeman, Aaron Grattafiori, Aarti Basant, Abhibha Gupta,
  Abhinav Khattar, Adi Renduchintala, Aditya Vavre, Akanksha Shukla, Akhiad
  Bercovich, Aleksander Ficek, Aleksandr Shaposhnikov, Alex Kondratenko,
  Alexander Bukharin, Alexandre Milesi, Ali Taghibakhshi, Alisa Liu, Amelia
  Barton, Ameya~Sunil Mahabaleshwarkar, and 294 others. 2025.
\newblock \href {https://arxiv.org/abs/2512.20848} {{Nemotron 3 Nano}: Open,
  efficient mixture-of-experts hybrid {Mamba}-{Transformer} model for agentic
  reasoning}.
\newblock \emph{Preprint}, arXiv:2512.20848.

\bibitem[{{OpenAI}(2026)}]{gpt56luna}
{OpenAI}. 2026.
\newblock \href {https://deploymentsafety.openai.com/gpt-5-6} {{GPT}-5.6
  {System Card}}.
\newblock OpenAI Deployment Safety Hub.

\bibitem[{Ouyang et~al.(2022)Ouyang, Wu, Jiang, Almeida, Wainwright, Mishkin,
  Zhang, Agarwal, Slama, Ray, Schulman, Hilton, Kelton, Miller, Simens, Askell,
  Welinder, Christiano, Leike, and Lowe}]{ouyang2022rlhf}
Long Ouyang, Jeffrey Wu, Xu~Jiang, Diogo Almeida, Carroll~L. Wainwright, Pamela
  Mishkin, Chong Zhang, Sandhini Agarwal, Katarina Slama, Alex Ray, John
  Schulman, Jacob Hilton, Fraser Kelton, Luke Miller, Maddie Simens, Amanda
  Askell, Peter Welinder, Paul~F. Christiano, Jan Leike, and Ryan Lowe. 2022.
\newblock \href
  {http://papers.nips.cc/paper\_files/paper/2022/hash/b1efde53be364a73914f58805a001731-Abstract-Conference.html}
  {Training language models to follow instructions with human feedback}.
\newblock In \emph{Advances in Neural Information Processing Systems 35: Annual
  Conference on Neural Information Processing Systems 2022, NeurIPS 2022, New
  Orleans, LA, USA, November 28 - December 9, 2022}.

\bibitem[{Pan et~al.(2026)Pan, Tao, Zhai, Zhang, Liu, Ding, Liu, and
  Wen}]{pan2026rlcsd}
Leyi Pan, Shuchang Tao, Yunpeng Zhai, Lingzhe Zhang, Zhaoyang Liu, Bolin Ding,
  Aiwei Liu, and Lijie Wen. 2026.
\newblock \href {https://arxiv.org/abs/2606.11709} {{RLCSD}: Reinforcement
  learning with contrastive on-policy self-distillation}.
\newblock \emph{Preprint}, arXiv:2606.11709.

\bibitem[{Penaloza et~al.(2026)Penaloza, Vattikonda, Gontier, Lacoste, Charlin,
  and Caccia}]{penaloza2026privilegedinformationdistillationlanguage}
Emiliano Penaloza, Dheeraj Vattikonda, Nicolas Gontier, Alexandre Lacoste,
  Laurent Charlin, and Massimo Caccia. 2026.
\newblock \href {https://arxiv.org/abs/2602.04942} {Privileged information
  distillation for language models}.
\newblock \emph{Preprint}, arXiv:2602.04942.

\bibitem[{Ross et~al.(2011)Ross, Gordon, and Bagnell}]{ross2011dagger}
St{\'{e}}phane Ross, Geoffrey~J. Gordon, and Drew Bagnell. 2011.
\newblock \href {http://proceedings.mlr.press/v15/ross11a/ross11a.pdf} {A
  reduction of imitation learning and structured prediction to no-regret online
  learning}.
\newblock In \emph{Proceedings of the Fourteenth International Conference on
  Artificial Intelligence and Statistics, {AISTATS} 2011, Fort Lauderdale, USA,
  April 11-13, 2011}, volume~15 of \emph{{JMLR} Proceedings}, pages 627--635.
  JMLR.org.

\bibitem[{Shao et~al.(2024)Shao, Wang, Zhu, Xu, Song, Bi, Zhang, Zhang, Li, Wu,
  and Guo}]{shao2024deepseekmath}
Zhihong Shao, Peiyi Wang, Qihao Zhu, Runxin Xu, Junxiao Song, Xiao Bi, Haowei
  Zhang, Mingchuan Zhang, Y.~K. Li, Y.~Wu, and Daya Guo. 2024.
\newblock \href {https://arxiv.org/abs/2402.03300} {{DeepSeekMath}: Pushing the
  limits of mathematical reasoning in open language models}.
\newblock \emph{Preprint}, arXiv:2402.03300.

\bibitem[{Sheng et~al.(2025)Sheng, Zhang, Ye, Wu, Zhang, Zhang, Peng, Lin, and
  Wu}]{sheng2025hybridflow}
Guangming Sheng, Chi Zhang, Zilingfeng Ye, Xibin Wu, Wang Zhang, Ru~Zhang,
  Yanghua Peng, Haibin Lin, and Chuan Wu. 2025.
\newblock \href {https://doi.org/10.1145/3689031.3696075} {{HybridFlow}: {A}
  flexible and efficient {RLHF} framework}.
\newblock In \emph{Proceedings of the Twentieth European Conference on Computer
  Systems, EuroSys 2025, Rotterdam, The Netherlands, 30 March 2025 - 3 April
  2025}, pages 1279--1297. {ACM}.

\bibitem[{Shin et~al.(2025)Shin, Ji, Gong, Kim, Choi, and
  Seo}]{shin2025generativepromptinternalization}
Haebin Shin, Lei Ji, Yeyun Gong, Sungdong Kim, Eunbi Choi, and Minjoon Seo.
  2025.
\newblock \href {https://doi.org/10.18653/v1/2025.naacl-long.376} {Generative
  prompt internalization}.
\newblock In \emph{Proceedings of the 2025 Conference of the Nations of the
  Americas Chapter of the Association for Computational Linguistics: Human
  Language Technologies (Volume 1: Long Papers)}, pages 7338--7363,
  Albuquerque, New Mexico. Association for Computational Linguistics.

\bibitem[{Shridhar et~al.(2023)Shridhar, Stolfo, and
  Sachan}]{shridhar-etal-2023-distilling}
Kumar Shridhar, Alessandro Stolfo, and Mrinmaya Sachan. 2023.
\newblock \href {https://doi.org/10.18653/v1/2023.findings-acl.441} {Distilling
  reasoning capabilities into smaller language models}.
\newblock In \emph{Findings of the Association for Computational Linguistics:
  ACL 2023}, pages 7059--7073, Toronto, Canada. Association for Computational
  Linguistics.

\bibitem[{Snell et~al.(2022)Snell, Klein, and
  Zhong}]{snell2022learningdistillingcontext}
Charlie Snell, Dan Klein, and Ruiqi Zhong. 2022.
\newblock \href {https://arxiv.org/abs/2209.15189} {Learning by distilling
  context}.
\newblock \emph{Preprint}, arXiv:2209.15189.

\bibitem[{Song and Zheng(2026)}]{song2026opdsurvey}
Mingyang Song and Mao Zheng. 2026.
\newblock \href {https://arxiv.org/abs/2604.00626} {A survey of on-policy
  distillation for large language models}.
\newblock \emph{Preprint}, arXiv:2604.00626.

\bibitem[{Sun et~al.(2026)Sun, Zheng, Song, Zhong, Cheng, Feng, Liu, Fang, and
  Wang}]{sun2026simctrecoveringlostsupervision}
Jie Sun, Mao Zheng, Mingyang Song, Qiyong Zhong, Yilin Cheng, Bichuan Feng,
  Pengfei Liu, Junfeng Fang, and Xiang Wang. 2026.
\newblock \href {https://arxiv.org/abs/2605.07711} {{SimCT}: Recovering lost
  supervision for cross-tokenizer on-policy distillation}.
\newblock \emph{Preprint}, arXiv:2605.07711.

\bibitem[{Vapnik and Vashist(2009)}]{vapnik2009lupi}
Vladimir Vapnik and Akshay Vashist. 2009.
\newblock \href {https://doi.org/10.1016/J.NEUNET.2009.06.042} {A new learning
  paradigm: Learning using privileged information}.
\newblock \emph{Neural Networks}, 22(5-6):544--557.

\bibitem[{Wang et~al.(2024)Wang, Ma, Zhang, Ni, Chandra, Guo, Ren, Arulraj, He,
  Jiang, Li, Ku, Wang, Zhuang, Fan, Yue, and Chen}]{wang2024mmlupro}
Yubo Wang, Xueguang Ma, Ge~Zhang, Yuansheng Ni, Abhranil Chandra, Shiguang Guo,
  Weiming Ren, Aaran Arulraj, Xuan He, Ziyan Jiang, Tianle Li, Max Ku, Kai
  Wang, Alex Zhuang, Rongqi Fan, Xiang Yue, and Wenhu Chen. 2024.
\newblock \href
  {http://papers.nips.cc/paper\_files/paper/2024/hash/ad236edc564f3e3156e1b2feafb99a24-Abstract-Datasets\_and\_Benchmarks\_Track.html}
  {{MMLU-Pro}: {A} more robust and challenging multi-task language
  understanding benchmark}.
\newblock In \emph{Advances in Neural Information Processing Systems 37: Annual
  Conference on Neural Information Processing Systems 2024, NeurIPS 2024,
  Vancouver, BC, Canada, December 10 - 15, 2024}.

\bibitem[{Wei et~al.(2022)Wei, Wang, Schuurmans, Bosma, Ichter, Xia, Chi, Le,
  and Zhou}]{wei2022cot}
Jason Wei, Xuezhi Wang, Dale Schuurmans, Maarten Bosma, Brian Ichter, Fei Xia,
  Ed~H. Chi, Quoc~V. Le, and Denny Zhou. 2022.
\newblock \href
  {http://papers.nips.cc/paper\_files/paper/2022/hash/9d5609613524ecf4f15af0f7b31abca4-Abstract-Conference.html}
  {Chain-of-thought prompting elicits reasoning in large language models}.
\newblock In \emph{Advances in Neural Information Processing Systems 35: Annual
  Conference on Neural Information Processing Systems 2022, NeurIPS 2022, New
  Orleans, LA, USA, November 28 - December 9, 2022}.

\bibitem[{Wu et~al.(2026)Wu, Han, and Cai}]{wu2026lightningopd}
Yecheng Wu, Song Han, and Han Cai. 2026.
\newblock \href {https://arxiv.org/abs/2604.13010} {Lightning {OPD}: Efficient
  post-training for large reasoning models with offline on-policy
  distillation}.
\newblock \emph{Preprint}, arXiv:2604.13010.

\bibitem[{Xiong et~al.(2026)Xiong, Shen, Gong, Cheng, Shen, Tao, Tan, Bai,
  Shang, and Wong}]{xiong2026ovdonpolicyverbaldistillation}
Jing Xiong, Hui Shen, Shansan Gong, Yuxin Cheng, Jianghan Shen, Chaofan Tao,
  Haochen Tan, Haoli Bai, Lifeng Shang, and Ngai Wong. 2026.
\newblock \href {https://arxiv.org/abs/2601.21968} {{OVD}: On-policy verbal
  distillation}.
\newblock \emph{Preprint}, arXiv:2601.21968.

\bibitem[{Xu et~al.(2025)Xu, Han, Wang, Le, Madeka, Li, Wang, Agarwal, Lee, and
  Pfister}]{xu2025skd}
Wenda Xu, Rujun Han, Zifeng Wang, Long~T. Le, Dhruv Madeka, Lei Li,
  William~Yang Wang, Rishabh Agarwal, Chen{-}Yu Lee, and Tomas Pfister. 2025.
\newblock \href {https://openreview.net/forum?id=EgJhwYR2tB} {Speculative
  knowledge distillation: Bridging the teacher-student gap through interleaved
  sampling}.
\newblock In \emph{The Thirteenth International Conference on Learning
  Representations, {ICLR} 2025, Singapore, April 24-28, 2025}. OpenReview.net.

\bibitem[{Yang et~al.(2025)Yang, Li, Yang, Zhang, Hui, Zheng, Yu, Gao, Huang,
  Lv, Zheng, Liu, Zhou, Huang, Hu, Ge, Wei, Lin, Tang, Yang, Tu, Zhang, Yang,
  Yang, Zhou, Zhou, Lin, Dang, Bao, Yang, Yu, Deng, Li, Xue, Li, Zhang, Wang,
  Zhu, Men, Gao, Liu, Luo, Li, Tang, Yin, Ren, Wang, Zhang, Ren, Fan, Su,
  Zhang, Zhang, Wan, Liu, Wang, Cui, Zhang, Zhou, and Qiu}]{qwen3report}
An~Yang, Anfeng Li, Baosong Yang, Beichen Zhang, Binyuan Hui, Bo~Zheng, Bowen
  Yu, Chang Gao, Chengen Huang, Chenxu Lv, Chujie Zheng, Dayiheng Liu, Fan
  Zhou, Fei Huang, Feng Hu, Hao Ge, Haoran Wei, Huan Lin, Jialong Tang, and 41
  others. 2025.
\newblock \href {https://arxiv.org/abs/2505.09388} {{Qwen3} technical report}.
\newblock \emph{Preprint}, arXiv:2505.09388.

\bibitem[{Yang et~al.(2026{\natexlab{a}})Yang, Qin, Si, Chen, Gu, Yao, Lin,
  Wang, Wang, and Duan}]{yang2026selfdistilled}
Chenxu Yang, Chuanyu Qin, Qingyi Si, Minghui Chen, Naibin Gu, Dingyu Yao, Zheng
  Lin, Weiping Wang, Jiaqi Wang, and Nan Duan. 2026{\natexlab{a}}.
\newblock \href {https://arxiv.org/abs/2604.03128} {Self-distilled {RLVR}}.
\newblock \emph{Preprint}, arXiv:2604.03128.

\bibitem[{Yang et~al.(2026{\natexlab{b}})Yang, Liu, Xie, Yang, Yang, and
  Lin}]{yang2026genopd}
Wenkai Yang, Weijie Liu, Ruobing Xie, Kai Yang, Saiyong Yang, and Yankai Lin.
  2026{\natexlab{b}}.
\newblock \href {https://arxiv.org/abs/2602.12125} {Learning beyond teacher:
  Generalized on-policy distillation with reward extrapolation}.
\newblock \emph{Preprint}, arXiv:2602.12125.

\bibitem[{Ye et~al.(2025{\natexlab{a}})Ye, Yang, Nan, Li, Zhang, Gui, Huang,
  Wang, Shi, and Fan}]{ye-etal-2025-analyzing}
Junjie Ye, Yuming Yang, Yang Nan, Shuo Li, Qi~Zhang, Tao Gui, Xuanjing Huang,
  Peng Wang, Zhongchao Shi, and Jianping Fan. 2025{\natexlab{a}}.
\newblock \href {https://doi.org/10.18653/v1/2025.emnlp-main.25} {Analyzing the
  effects of supervised fine-tuning on model knowledge from token and parameter
  levels}.
\newblock In \emph{Proceedings of the 2025 Conference on Empirical Methods in
  Natural Language Processing}, pages 471--513, Suzhou, China. Association for
  Computational Linguistics.

\bibitem[{Ye et~al.(2026{\natexlab{a}})Ye, Dong, Chi, Wu, Huang, and
  Wei}]{ye2026blackboxonpolicydistillationlarge}
Tianzhu Ye, Li~Dong, Zewen Chi, Xun Wu, Shaohan Huang, and Furu Wei.
  2026{\natexlab{a}}.
\newblock \href {https://arxiv.org/abs/2511.10643} {Black-box on-policy
  distillation of large language models}.
\newblock \emph{Preprint}, arXiv:2511.10643.

\bibitem[{Ye et~al.(2026{\natexlab{b}})Ye, Dong, Wu, Huang, and
  Wei}]{ye2026onpolicycontextdistillationlanguage}
Tianzhu Ye, Li~Dong, Xun Wu, Shaohan Huang, and Furu Wei. 2026{\natexlab{b}}.
\newblock \href {https://arxiv.org/abs/2602.12275} {On-policy context
  distillation for language models}.
\newblock \emph{Preprint}, arXiv:2602.12275.

\bibitem[{Ye et~al.(2025{\natexlab{b}})Ye, Huang, Xiao, Chern, Xia, and
  Liu}]{ye2025limoreasoning}
Yixin Ye, Zhen Huang, Yang Xiao, Ethan Chern, Shijie Xia, and Pengfei Liu.
  2025{\natexlab{b}}.
\newblock \href {https://openreview.net/forum?id=T2TZ0RY4Zk} {{LIMO}: Less is
  more for reasoning}.
\newblock In \emph{Second Conference on Language Modeling}.

\bibitem[{Yu et~al.(2025)Yu, Zhang, Zhu, Yuan, Zuo, Yue, Dai, Fan, Liu, Liu,
  Liu, Liu, Lin, Lin, Ma, Sheng, Tong, Zhang, Zhang, Zhang, Zhang, Zhu, Zhu,
  Chen, Chen, Wang, Yu, Song, Wei, Zhou, Liu, Ma, Zhang, Yan, Wu, and
  Wang}]{yu2025dapo}
Qiying Yu, Zheng Zhang, Ruofei Zhu, Yufeng Yuan, Xiaochen Zuo, Yu~Yue, Weinan
  Dai, Tiantian Fan, Gaohong Liu, Juncai Liu, Lingjun Liu, Xin Liu, Haibin Lin,
  Zhiqi Lin, Bole Ma, Guangming Sheng, Yuxuan Tong, Chi Zhang, Mofan Zhang, and
  17 others. 2025.
\newblock \href
  {http://papers.nips.cc/paper\_files/paper/2025/hash/a4277440d50f1f15d2cb4c14f7e0c0d2-Abstract-Conference.html}
  {{DAPO:} an open-source {LLM} reinforcement learning system at scale}.
\newblock In \emph{Advances in Neural Information Processing Systems 38: Annual
  Conference on Neural Information Processing Systems 2025, NeurIPS 2025, San
  Diego, CA, USA, December 2-7, 2025 / Mexico City, Mexico, November 30 -
  December 5, 2025}.

\bibitem[{Zhang et~al.(2025)Zhang, Li, Song, Ye, Gao, and
  Hu}]{zhang-etal-2025-towards-law}
Chen Zhang, Qiuchi Li, Dawei Song, Zheyu Ye, Yan Gao, and Yao Hu. 2025.
\newblock \href {https://doi.org/10.18653/v1/2025.acl-long.1097} {Towards the
  law of capacity gap in distilling language models}.
\newblock In \emph{Proceedings of the 63rd Annual Meeting of the Association
  for Computational Linguistics (Volume 1: Long Papers)}, pages 22504--22528,
  Vienna, Austria. Association for Computational Linguistics.

\bibitem[{Zhang et~al.(2023)Zhang, Yang, Liu, Wang, Xian, Wang, and
  Song}]{zhang-etal-2023-lifting}
Chen Zhang, Yang Yang, Jiahao Liu, Jingang Wang, Yunsen Xian, Benyou Wang, and
  Dawei Song. 2023.
\newblock \href {https://doi.org/10.18653/v1/2023.acl-long.249} {Lifting the
  curse of capacity gap in distilling language models}.
\newblock In \emph{Proceedings of the 61st Annual Meeting of the Association
  for Computational Linguistics (Volume 1: Long Papers)}, pages 4535--4553,
  Toronto, Canada. Association for Computational Linguistics.

\bibitem[{Zhang et~al.(2024)Zhang, Zhang, Sun, Chen, and
  Xu}]{zhang-etal-2024-dual}
Songming Zhang, Xue Zhang, Zengkui Sun, Yufeng Chen, and Jinan Xu. 2024.
\newblock \href {https://doi.org/10.18653/v1/2024.emnlp-main.1010} {Dual-space
  knowledge distillation for large language models}.
\newblock In \emph{Proceedings of the 2024 Conference on Empirical Methods in
  Natural Language Processing}, pages 18164--18181, Miami, Florida, USA.
  Association for Computational Linguistics.

\bibitem[{Zhang et~al.(2026)Zhang, Ding, Pan, Yang, Kang, Xiong, and
  Gu}]{zhang2026opsdlonpolicyselfdistillationlongcontext}
Xinsen Zhang, Zhenkai Ding, Tianjun Pan, Run Yang, Chun Kang, Xue Xiong, and
  Jingnan Gu. 2026.
\newblock \href {https://arxiv.org/abs/2604.17535} {{OPSDL}: On-policy
  self-distillation for long-context language models}.
\newblock \emph{Preprint}, arXiv:2604.17535.

\bibitem[{Zhao et~al.(2026)Zhao, Xie, Liu, Huang, Pang, Chen, and
  Grover}]{zhao2026opsd}
Siyan Zhao, Zhihui Xie, Mengchen Liu, Jing Huang, Guan Pang, Feiyu Chen, and
  Aditya Grover. 2026.
\newblock \href {https://arxiv.org/abs/2601.18734} {Self-distilled reasoner:
  On-policy self-distillation for large language models}.
\newblock \emph{Preprint}, arXiv:2601.18734.

\bibitem[{Zhou et~al.(2026)Zhou, Zhang, Wu, Wang, Peng, Liu, Fan, and
  Zhao}]{zhou2026omniopd}
Yuhang Zhou, Lizhu Zhang, Yifan Wu, Mingyi Wang, Bo~Peng, Jiayi Liu, Xiangjun
  Fan, and Zhuokai Zhao. 2026.
\newblock \href {https://arxiv.org/abs/2606.01476} {{OmniOPD}: Logit-free
  on-policy distillation via speculative verification}.
\newblock \emph{Preprint}, arXiv:2606.01476.

\end{thebibliography}

\appendix

\section{Experimental Details}
\label{app:exp_details}

\paragraph{Teacher traces.}
Teacher traces are generated once, before training, with the settings in
Table~\ref{tab:trace_config} and the prompts in Appendix~\ref{app:prompts}.
For a closed teacher that hides its reasoning, the trace is built from the
response it returns: we enclose the derivation of the response in a
reasoning block (\texttt{<think>} and \texttt{</think>}) and place its
boxed final answer after the block, which is the format of the students'
own responses. Within a teacher--student pair, BSD and SFT use the same traces, BSD
as privileged context for the self-teacher and SFT as training targets.

\begin{table*}[t]
\begin{center}
\small
\begin{tabular}{lccc}
\toprule
Parameter & \texttt{Qwen3-32B} & \texttt{Nemotron-3-30B-A3B} & \texttt{GPT-5.6 Luna} \\
\midrule
Access & Local weights & Local weights & API \\
Thinking mode & Enabled & Enabled & Reasoning effort medium \\
Temperature & 0.6 & 1.0 & Provider default \\
Top-p & 0.95 & 1.0 & Provider default \\
Max new tokens & 8{,}192 & 8{,}192 & 8{,}192 \\
Traces per problem & 1 & 1 & 1 \\
\bottomrule
\end{tabular}
\end{center}
\caption{Generation settings of the teacher traces. The open teachers follow
their recommended sampling settings.}
\label{tab:trace_config}
\end{table*}

\paragraph{Training.}
Tables~\ref{tab:train_config} and~\ref{tab:bsd_config} give the training
configuration. All methods train LoRA adapters~\citep{hu2022lora}, and the
learning rate is warmed up linearly over the first 10 updates and is
constant afterwards. Training sets are identical for all methods within a
teacher--student pair. For BSD, the student's score $p_t$ uses its current
log-probabilities, and the rollout--trainer gap is corrected with
token-level truncated importance sampling~\citep{yu2025dapo}.
During training, BSD adds one forward pass of the self-teacher over the
student's rollouts. This pass takes 7 to 18 seconds per update, depending
on the length of the trace, which is under 10\% of the update time.
The limit of 2{,}048 tokens on the prompt applies to the student's
prompt, which contains only the question. The self-teacher's prompt also
contains the trace and has a separate limit of 12{,}288 tokens. The
longest question and trace together take 8{,}950 tokens, so no trace is
truncated in any run; the hint and guide of Section~\ref{sec:ablation}
are shorter than the full trace.

\begin{table*}[t]
\begin{center}
\small
\begin{tabular}{lccccc}
\toprule
Parameter & BSD & OPD & xOPD & ROPD & SFT \\
\midrule
Learning rate & $1 \times 10^{-6}$ & $1 \times 10^{-6}$ & $1 \times 10^{-6}$ & $1 \times 10^{-6}$ & $5 \times 10^{-6}$ \\
Training updates & 150 & 150 & 150 & 150 & 150 \\
LoRA rank ($r$) & 32 & 32 & 32 & 32 & 32 \\
LoRA alpha ($\alpha$) & 64 & 64 & 64 & 64 & 64 \\
Checkpoint interval (updates) & 25 & 25 & 25 & 25 & 25 \\
Max prompt length (student) & 2{,}048 & 2{,}048 & 2{,}048 & 2{,}048 & -- \\
Max prompt length (self-teacher) & 12{,}288 & -- & -- & -- & -- \\
Max response length & 12{,}288 & 12{,}288 & 12{,}288 & 8{,}192 & -- \\
Problems per update & 16 & 16 & 16 & 16 & 16 \\
Rollouts per problem & 8 & 1 & 8 & 8 & -- \\
Sampling temperature & 1.0 & 1.0 & 1.0 & 1.0 & -- \\
\bottomrule
\end{tabular}
\end{center}
\caption{Training configuration of BSD and the baselines. The rollout
setting of OPD follows its original implementation, and the maximum
response length of ROPD follows its released code.}
\label{tab:train_config}
\end{table*}

\begin{table}[t]
\begin{center}
\small
\begin{tabular}{@{}>{\raggedright\arraybackslash}p{0.44\columnwidth}>{\raggedright\arraybackslash}p{0.5\columnwidth}@{}}
\toprule
Parameter & Value \\
\midrule
Weight coefficient $\lambda$ & $0.5$, decayed linearly to $0$ over training \\
Clipping range $\epsilon$ of the weight & $0.2$ \\
Self-teacher synchronization & Every 10 updates \\
Policy clipping thresholds & $0.2$ (lower), $0.28$ (upper) \\
Loss aggregation & Token mean \\
Importance sampling & Token-level, truncated \\
KL penalty & None \\
Entropy regularization & None \\
\bottomrule
\end{tabular}
\end{center}
\caption{Settings specific to BSD.}
\label{tab:bsd_config}
\end{table}

\paragraph{Compute.}
In total, the training, evaluation, and ablation runs reported in this
paper take about 1{,}450 GPU-hours on NVIDIA H100 and H200 GPUs.
Table~\ref{tab:train_time} reports the GPU-hours of training with each
method for the \texttt{Qwen3-4B} student, measured on NVIDIA H200 GPUs.
The values cover the 150 updates and exclude checkpoint saving,
checkpoint evaluation, and the generation of the teacher traces before
training. OPD and xOPD keep the teacher on one additional GPU during
training, which is counted, and ROPD queries the teacher through its API,
whose compute is not counted.

\begin{table}[t]
\begin{center}
\small
\begin{tabular}{llc}
\toprule
Teacher & Method & GPU-hours \\
\midrule
\multirow{3}{*}{\texttt{Qwen3-32B}} & SFT & 0.8 \\
 & OPD & 29.4 \\
 & BSD & 31.4 \\
\midrule
\multirow{2}{*}{\texttt{Nemotron-3-30B-A3B}} & xOPD & 70.8 \\
 & BSD & 48.7 \\
\midrule
\multirow{2}{*}{\texttt{GPT-5.6 Luna}} & ROPD & 83.0 \\
 & BSD & 33.3 \\
\bottomrule
\end{tabular}
\end{center}
\caption{GPU-hours of training the \texttt{Qwen3-4B} student for 150
updates with each method. The value of SFT is measured with the
\texttt{Qwen3-32B} traces.}
\label{tab:train_time}
\end{table}

\paragraph{Evaluation.}
Table~\ref{tab:eval_config} gives the evaluation parameters. Checkpoints
are selected on the development set, and the earliest checkpoint is taken
in case of a tie. Each of the three benchmarks contains 30 problems. Final
answers are checked with
Math-Verify.\footnote{\url{https://github.com/huggingface/Math-Verify}}

\begin{table}[t]
\begin{center}
\small
\begin{tabular}{ll}
\toprule
Parameter & Value \\
\midrule
Samples per problem & 4 \\
Seeds & $\{0, 1, 2, 3\}$ \\
Thinking mode & Enabled \\
Temperature & 1.0 \\
Top-p & 1.0 \\
Max new tokens & 32{,}768 \\
Checkpoint selection & 128 development problems \\
\bottomrule
\end{tabular}
\end{center}
\caption{Evaluation parameters for the students on the three benchmarks.}
\label{tab:eval_config}
\end{table}

\paragraph{Guidance measurement.}
For Table~\ref{tab:guidance}, each student generates six rollouts for each
of 24 hard problems shared by all students before training, and each rollout is scored
under one verified trace from each of the three teachers. We write
$\bar\delta$ for the mean of the signed guidance $\delta_t$ over all
tokens of the unsuccessful rollouts, and Table~\ref{tab:guidance} reports its absolute
value $|\bar\delta|$. The mean is negative for every teacher--student
pair, so the self-teacher assigns on average a lower probability to the
student's tokens than the student does. This quantity differs from the
mean absolute guidance $\overline{|\delta|}$, in which positive and
negative values do not cancel; it is 0.163, 0.191, 0.202, and 0.110 for
\texttt{Qwen3-0.6B}, \texttt{Qwen3-1.7B}, \texttt{Qwen3-4B}, and
\texttt{Qwen3-4B-Instruct}, respectively, so the last student has the
smallest value under both statistics. The correlation $\rho_H$ is computed within
each of the same rollouts and averaged over them.

\paragraph{Teacher evaluation.}
Table~\ref{tab:teacher_scores} reports Avg@4 of the external teachers on
the three benchmarks, with four samples per problem and seeds $\{0,1,2,3\}$.
\texttt{Qwen3-32B} and \texttt{Nemotron-3-30B-A3B} are evaluated under the students' protocol:
thinking enabled, temperature 1.0, top-p 1.0, and a maximum generation
length of 32{,}768 tokens. \texttt{GPT-5.6 Luna} is evaluated through its API with
reasoning effort medium, a limit of 32{,}768 completion tokens, and the
provider's default sampling parameters; its reasoning is not returned.

\begin{table}[t]
\begin{center}
\small
\setlength{\tabcolsep}{5pt}
\resizebox{\columnwidth}{!}{%
\begin{tabular}{lcccc}
\toprule
Teacher & AIME24 & AIME25 & HMMT25 & Avg \\
\midrule
\texttt{Qwen3-32B} & 80.0 & 75.0 & 52.5 & 69.2 \\
\texttt{Nemotron-3-30B-A3B} & 80.0 & 70.0 & 55.8 & 68.6 \\
\texttt{GPT-5.6 Luna} & 88.3 & 87.5 & 85.8 & 87.2 \\
\bottomrule
\end{tabular}}
\end{center}
\caption{Avg@4 of the external teachers on the evaluation benchmarks and
their mean (Avg).}
\label{tab:teacher_scores}
\end{table}

\paragraph{Fine-tuning of \texttt{Qwen3-4B-Instruct}.}
The released \texttt{Qwen3-4B-Instruct-2507} does not produce a reasoning block, so
we fine-tune it on 1{,}000 \texttt{Nemotron-3-30B-A3B} traces before using it as a student
(Section~\ref{sec:exp_setup}). Table~\ref{tab:instruct_base} reports its
scores before and after this step. The released model is evaluated with
its plain prompt and no reasoning block; the fine-tuned model is the base
model of the \texttt{Qwen3-4B-Instruct} block in Table~\ref{tab:main}.

\begin{table}[t]
\begin{center}
\small
\setlength{\tabcolsep}{5pt}
\begin{tabular}{lcccc}
\toprule
Model & AIME24 & AIME25 & HMMT25 & Avg \\
\midrule
Released & 60.8 & 49.2 & 29.2 & 46.4 \\
Fine-tuned & 56.7 & 49.2 & 27.5 & 44.4 \\
\bottomrule
\end{tabular}
\end{center}
\caption{Avg@4 of \texttt{Qwen3-4B-Instruct-2507} before and after the fine-tuning
that enables its reasoning block, and their mean (Avg).}
\label{tab:instruct_base}
\end{table}

\paragraph{Licenses.}
Our code builds on verl, which is released under the Apache License 2.0, and
we release it under the same license. The \texttt{Qwen3} models are released
under the Apache License 2.0 and \texttt{Nemotron-3-30B-A3B} under the NVIDIA
Nemotron Open Model License; the closed teacher is accessed under the terms
of its API. OpenThoughts is released under the Apache License 2.0, and
HMMT 2025 under CC BY-NC-SA 4.0. We use all models and datasets for research
only and redistribute none of them.

\section{Prompt Templates}
\label{app:prompts}

All prompts are given as the user message of the model's chat template.
For \texttt{Qwen3-4B}, \texttt{Qwen3-1.7B}, and \texttt{Qwen3-0.6B}, the student is prompted with the question:

\begin{quote}
\small\raggedright
Problem: \{question $x$\}

Please reason step by step, and put your final answer within
\texttt{\textbackslash boxed\{\}}.
\end{quote}

The self-teacher receives the same question together with the external
teacher's trace:

\begin{quote}
\small\raggedright
Problem: \{question $x$\}

Here is a reference solution to this problem:

\texttt{<reference\_solution>}\\
\{trace $z$\}\\
\texttt{</reference\_solution>}

After reading the reference solution above, make sure you truly understand
the reasoning behind each step --- do not copy or paraphrase it. Now, using
your own words and independent reasoning, derive the same final answer to
the problem above. Think step by step, explore different approaches, and
don't be afraid to backtrack or reconsider if something doesn't work out:

Please reason step by step, and put your final answer within
\texttt{\textbackslash boxed\{\}}.
\end{quote}

For \texttt{Qwen3-4B-Instruct}, the chat template opens the reasoning
block (\texttt{<think>}) at the start of the response, and the last line
of both prompts is additionally replaced by:

\begin{quote}
\small\raggedright
Think through the problem step by step inside \texttt{<think>} and
\texttt{</think>} tags. After \texttt{</think>}, give your final solution
and put the final answer within \texttt{\textbackslash boxed\{\}}.
\end{quote}

The token-level guidance is computed on the student's original response
tokens appended after each prompt; responses are not decoded and
re-tokenized.

\paragraph{Teacher traces.}
All three external teachers generate their traces from the same prompt:

\begin{quote}
\small\raggedright
Problem: \{question $x$\}

Please reason step by step, and put your final answer within
\texttt{\textbackslash boxed\{\}}.
\end{quote}

\paragraph{RLSD ablation.}
For RLSD (Section~\ref{sec:ablation}), the self-teacher receives the
ground-truth answer in place of the trace:

\begin{quote}
\small\raggedright
Problem: \{question $x$\}

Here is a verified correct final answer to this problem:

\texttt{<reference\_solution>}\\
\{answer\}\\
\texttt{</reference\_solution>}

After reading the reference solution above, make sure you truly understand
the reasoning behind each step --- do not copy or paraphrase it. Now, using
your own words and independent reasoning, derive the same final answer to
the problem above. Think step by step, explore different approaches, and
don't be afraid to backtrack or reconsider if something doesn't work out:

Please reason step by step, and put your final answer within
\texttt{\textbackslash boxed\{\}}.
\end{quote}

\paragraph{Guide and hint ablations.}
For the ablation on the form of the privileged context
(Section~\ref{sec:ablation}), the external teacher rewrites its trace as
a guide or as a hint. The guide is generated with:

\begin{quote}
\small\raggedright
You are helping a student who is stuck on a math problem. Below are the
problem and a worked reasoning trace written by a stronger solver.

Problem:\\
\{question $x$\}

Worked reasoning:\\
\{trace $z$\}

Write a step-by-step guide for the student: a numbered list of the steps
of the solution, in order, at most 8 steps and under 150 words in total.
Rules: each step says what to do and why (the idea, the quantity to set
up, the case split), without carrying out the computation; do not state
the final answer; do not give intermediate numerical results; do not
restate the problem; plain text, no markdown formatting. Output only the
numbered list.
\end{quote}

The hint is generated with the same opening, problem, and worked
reasoning, followed by this instruction:

\begin{quote}
\small\raggedright
Write a hint of at most three sentences (under 80 words) that points the
student to the key idea or the first decisive step. Rules: do not state
the final answer; do not give numerical intermediate results; do not
write out the solution; name the method or the observation that unlocks
the problem. Output only the hint.
\end{quote}

The self-teacher then receives the guide, with the verified final answer
appended, in place of the reference solution:

\begin{quote}
\small\raggedright
Problem: \{question $x$\}

Here is a step-by-step guide to this problem, followed by its verified
final answer:

\texttt{<guide>}\\
\{guide\}\\
\texttt{</guide>}

After reading the guide above, make sure you truly understand why each
step is needed --- do not copy or paraphrase it. Now, using your own words
and independent reasoning, carry out the steps and derive the same final
answer to the problem above. Think step by step, explore different
approaches, and don't be afraid to backtrack or reconsider if something
doesn't work out:

Please reason step by step, and put your final answer within
\texttt{\textbackslash boxed\{\}}.
\end{quote}

The self-teacher prompt for the hint is:

\begin{quote}
\small\raggedright
Problem: \{question $x$\}

Here is a hint for this problem, followed by its verified final answer:

\texttt{<hint>}\\
\{hint\}\\
\texttt{</hint>}

After reading the hint above, make sure you truly understand the idea it
points to --- do not copy or paraphrase it. Now, using your own words and
independent reasoning, work out the solution and derive the same final
answer to the problem above. Think step by step, explore different
approaches, and don't be afraid to backtrack or reconsider if something
doesn't work out:

Please reason step by step, and put your final answer within
\texttt{\textbackslash boxed\{\}}.
\end{quote}

\section{Proofs}
\label{app:proofs}

Throughout, expectations are over prompts $x$, traces $z$, and
rollouts $y \sim \pi_{\theta_{\mathrm{old}}}(\cdot \mid x)$, where
$\theta_{\mathrm{old}}$ is the policy that sampled the rollouts and
$\bar\theta$ is the frozen copy of the student that serves as the
self-teacher (Section~\ref{sec:selfteacher}), so that $p_t =
\pi_{\theta_{\mathrm{old}}}(y_t \mid x, y_{<t})$ and $q_t =
\pi_{\bar\theta}(y_t \mid x, z, y_{<t})$ in Eq.~\ref{eq:delta}; the
response mask is left implicit. The quantities $A$, $\delta_t$, and
$w_t$ carry no gradient, so the statements below do not depend on how
often the self-teacher is synchronized.

\subsection{Small-Gap Decomposition}
\label{app:smallgap_statement}

Reverse-KL OPD trains the student with the token-level advantage
$\hat A_t^{\mathrm{OPD}} = \log \pi^{*}(y_t \mid s_t) - \log \pi_\theta(y_t
\mid s_t)$, where $\pi^{*}$ is the teacher and $s_t = (x, y_{<t})$; with a
reference policy $\pi_{\mathrm{ref}}$ it is equivalent to KL-regularized RL
with the token reward $r_t = \log \pi^{*}(y_t \mid s_t) - \allowbreak
\log \pi_{\mathrm{ref}}(y_t \mid s_t)$~\citep{agarwal2024gkd, yang2026genopd}.
Taking the self-teacher as $\pi^{*}$ and the unconditioned student as
$\pi_{\mathrm{ref}}$ gives $\log q_t - \log p_t$, whose stop-gradient
value is $\delta_t$, so $r_t^{\mathrm{OPD},z} = \delta_t$. Writing $\Phi_{\lambda,\epsilon}(u) =
(1-\lambda) + \lambda\, \clipe{e^{-u}}$, Eqs.~\ref{eq:routing} and~\ref{eq:weight} read
\begin{equation}
  \tilde A_t \;=\; A \cdot \Phi_{\lambda,\epsilon}(\delta_t)\,
  \mathbf{1}_{\{A<0\}} \;+\; A\, \mathbf{1}_{\{A \ge 0\}},
  \label{eq:phi}
\end{equation}
and since $\Phi_{\lambda,\epsilon} > 0$, $\sign(\tilde A_t) = \sign(A)$,
whereas the sign of $\hat A_t^{\mathrm{OPD}}$ follows $\delta_t$.

\begin{proposition}[Small-gap decomposition]
\label{prop:smallgap}
Ignoring clipping and evaluating the update at the policy that sampled
the rollouts, $\theta = \theta_{\mathrm{old}}$, for $|\delta_t| \ll 1$ the
BSD policy gradient
satisfies
\begin{multline}
  g_{\mathrm{BSD}}
  \;=\;
  g_{\mathrm{GRPO}}
  \;+\; O(\delta_t^2) \\
  \;+\;
  \lambda\,
  \mathbb{E}\!\left[
  \textstyle\sum_t \mathbf{1}_{\{A<0\}}\, |A|\; \delta_t\,
  \nabla_\theta \log \pi_\theta(y_t \mid s_t)
  \right].
\end{multline}
\end{proposition}

This decomposition is an interpretation and does not guarantee stability
or good credit assignment. It holds for any privileged context $z$. With
$z$ the ground-truth answer it describes the RLSD variant of
Section~\ref{sec:ablation}, which also restricts the weight to
unsuccessful rollouts; with a teacher trace, $r_t^{\mathrm{OPD},z}$ carries
information from another model's reasoning.

\subsection{Proof of Proposition~\ref{prop:smallgap}}
\label{app:proof_smallgap}

The weight of Eq.~\ref{eq:weight} applies only to tokens with $A < 0$;
tokens with $A \ge 0$ contribute their GRPO term unchanged. On $A < 0$ the BSD
token advantage is
\begin{equation}
  \tilde A_t = A\big[(1-\lambda) + \lambda e^{-\delta_t}\big],
\end{equation}
where clipping is ignored (it is inactive for $|\delta_t|$ small, since the
clip interval $[1-\epsilon, 1+\epsilon]$ contains a neighborhood of $1$).
Expanding $e^{-\delta_t} = 1 - \delta_t + O(\delta_t^2)$ gives
\begin{equation}
\begin{split}
  \tilde A_t
  &= A\big[1 - \lambda\,\delta_t\big] + O\!\big(|A|\,\delta_t^2\big) \\
  &= A - \lambda\,A\,\delta_t + O\!\big(|A|\,\delta_t^2\big).
\end{split}
\end{equation}
Substituting into the policy gradient evaluated at the rollout policy,
$g = \mathbb{E}[\sum_t \tilde A_t \nabla_\theta \log \pi_\theta(y_t \mid
s_t)]\big|_{\theta = \theta_{\mathrm{old}}}$, where the PPO likelihood ratio
equals one and its clipping is inactive, and writing $-A = |A|$ on the
reweighted tokens ($A < 0$) yields
\begin{multline}
  g_{\mathrm{BSD}}
  = \underbrace{\mathbb{E}\Big[\textstyle\sum_t A\, \nabla_\theta \log \pi_\theta\Big]}_{g_{\mathrm{GRPO}}}
  + O(\delta_t^2) \\
  + \lambda\, \mathbb{E}\Big[\textstyle\sum_t \mathbf{1}_{\{A<0\}}\, |A|\, \delta_t\,
    \nabla_\theta \log \pi_\theta\Big].
\end{multline}
Since $\delta_t = r_t^{\mathrm{OPD},z}$, the
correction is an outcome-weighted OPD update restricted to unsuccessful
rollouts.
\qed

\subsection{Weighting Successful Rollouts}
\label{app:success_branch}

Proposition~\ref{prop:smallgap} concerns unsuccessful rollouts only. Two
different weights can be applied to rollouts with $A > 0$, and they act in
opposite directions. The first is the weight of RLSD
(Eq.~\ref{eq:rlsd}), $(1-\lambda) + \lambda\,\clipe{e^{+\delta_t}}$,
which is the one used by the BSD (w/ succ.) variant of
Section~\ref{sec:ablation}. Under the conditions of
Proposition~\ref{prop:smallgap}, it gives
\begin{equation}
\begin{split}
  \tilde A_t
  &= A\big[(1-\lambda) + \lambda e^{\delta_t}\big] \\
  &= A + \lambda\,|A|\,\delta_t + O\!\big(|A|\,\delta_t^2\big),
\end{split}
\end{equation}
so the correction has the same form as on unsuccessful rollouts. The
update of BSD (w/ succ.) is then the GRPO update plus an outcome-weighted
OPD correction on all rollouts, which also pulls successful rollouts
toward the self-teacher. The second applies the weight of
Eq.~\ref{eq:weight}, with $e^{-\delta_t}$, unchanged to successful
rollouts. It gives $\tilde A_t = A - \lambda\,|A|\,\delta_t +
O(|A|\,\delta_t^2)$, whose correction has the opposite sign and pushes
the policy away from the self-teacher on successful rollouts, rewarding
divergence as in RLRT~\citep{kim2026rlrt}. We do not evaluate this second
weight.

BSD uses neither weight. A successful rollout is a solution that the
student found without the trace, and BSD reinforces it with its GRPO
advantage alone, without moving it toward or away from the self-teacher.
The expansion above describes what the first weight adds but does not by
itself predict its effect on accuracy; the support for restricting the
weight to unsuccessful rollouts is empirical, as BSD (w/ succ.) scores
61.4 against 63.9 for BSD (Section~\ref{sec:ablation}).

\section{Ablation Details}
\label{app:ablation_scores}

Table~\ref{tab:ablation} gives the scores of
Figure~\ref{fig:ablation} on each benchmark, with the training step
of each selected checkpoint and the magnitude of the mean token-level
guidance of each form of context.

\paragraph{Training settings of the ablation variants.}
GRPO, RLSD, and OPSD use the training problems, optimizer, learning rate,
rollout budget (16 problems with 8 rollouts each), number of updates, and
length limits of BSD (Table~\ref{tab:train_config}), and
Table~\ref{tab:ablation_config} lists where they differ. GRPO trains on
the group-normalized verifier reward alone. RLSD applies the weight and
schedule of BSD to unsuccessful rollouts, with a self-teacher that sees
only the ground-truth answer. OPSD~\citep{zhao2026opsd} conditions a
frozen copy of the base model on the full trace, following its original
implementation, and uses the clipped
log-probability difference between this self-teacher and the student as a
token-level reward on every rollout, without the verifier reward.
BSD-hint, BSD-guide, and BSD (w/ succ.) share the settings of BSD-trace
except for the privileged context or, for BSD (w/ succ.), the weight on
successful rollouts, which follows Eq.~\ref{eq:rlsd}.

\begin{table}[t]
\begin{center}
\small
\setlength{\tabcolsep}{3.5pt}
\resizebox{\columnwidth}{!}{%
\begin{tabular}{llll}
\toprule
Setting & GRPO & RLSD & OPSD \\
\midrule
Privileged context & None & Answer & Full trace \\
Self-teacher & None & Synced every 10 updates & Frozen base model \\
Verifier reward & Yes & Yes & No \\
Token-level signal & None & Weight, $\lambda$: $0.5 \to 0$, $\epsilon = 0.2$ & Log-ratio, clipped at $\pm 2$ \\
Rollouts with the signal & -- & Unsuccessful & All \\
\bottomrule
\end{tabular}}
\end{center}
\caption{Settings in which GRPO, RLSD, and OPSD differ in the ablation. All other
settings follow Table~\ref{tab:train_config}.}
\label{tab:ablation_config}
\end{table}

\begin{figure}[t]
  \centering
  \includegraphics[width=\linewidth]{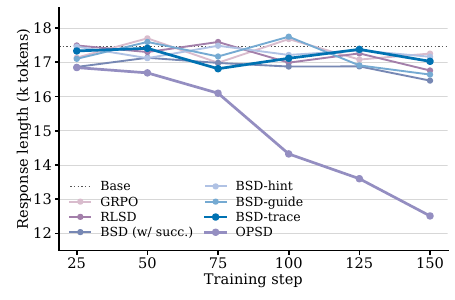}
  \caption{Mean response length on the three benchmarks at every saved
  checkpoint for the variants of Figure~\ref{fig:ablation}, with the
  \texttt{Qwen3-4B} student and the \texttt{Qwen3-32B} external teacher.
  The dotted line marks the base model.}
  \label{fig:ablation_length}
\end{figure}

Figure~\ref{fig:ablation_length} shows the mean response length of each
variant on the three benchmarks at every saved checkpoint. The responses
of OPSD shorten from 16.8k tokens at step 25 to 12.5k at step 150,
against 17.5k for the base model, and its Avg@4 falls from 63.3 to 55.8
over the same steps. The other variants stay between 16.5k and 17.7k
tokens.

\begin{table}[t]
\begin{center}
\small
\setlength{\tabcolsep}{3.5pt}
\resizebox{\columnwidth}{!}{%
\begin{tabular}{lcccccc}
\toprule
Method & AIME24 & AIME25 & HMMT25 & Avg & Step & $|\bar\delta|$ \\
\midrule
Base & 72.5 & 62.5 & 42.5 & 59.2 & -- & -- \\
\midrule
OPSD & 70.8 & 66.7 & 41.7 & 59.7 & 50 & -- \\
GRPO & 72.5 & 64.2 & 45.0 & 60.6 & 50 & -- \\
RLSD & 73.3 & \textbf{69.2} & 42.5 & 61.7 & 125 & $0.055$ \\
\midrule
BSD (w/ succ.) & 75.8 & 65.0 & 43.3 & 61.4 & 125 & $0.127$ \\
BSD-hint & 71.7 & 68.3 & 45.0 & 61.7 & 150 & $0.062$ \\
BSD-guide & 73.3 & 67.5 & \textbf{46.7} & 62.5 & 75 & $0.076$ \\
\rowcolor{bsdrow} BSD-trace & \textbf{79.2} & 68.3 & 44.2 & \textbf{63.9} & 125 & $0.127$ \\
\bottomrule
\end{tabular}}
\end{center}
\caption{Scores of the ablation in Figure~\ref{fig:ablation}, with the
\texttt{Qwen3-4B} student and the \texttt{Qwen3-32B} training set. Protocol as
in Table~\ref{tab:main}; Step is
the training step of the selected checkpoint, and $|\bar\delta|$ is the
magnitude of the mean token-level guidance of each form of context, measured before
training on the same unsuccessful rollouts (Section~\ref{sec:anatomy}).}
\label{tab:ablation}
\end{table}

\begin{figure}[t]
  \centering
  \includegraphics[width=\columnwidth]{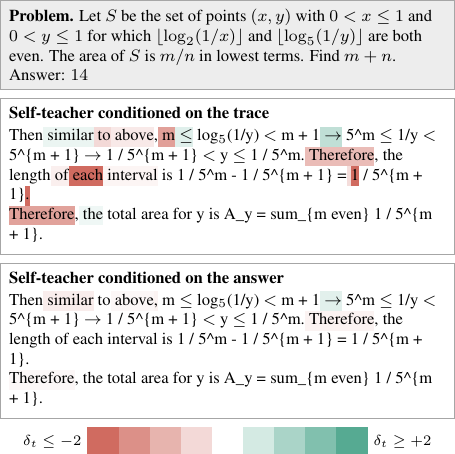}
  \caption{Token-level guidance $\delta_t$ on one passage of an
  unsuccessful rollout, with the self-teacher conditioned on the
  \texttt{Qwen3-32B} trace (top) or on the answer alone (bottom). Red
  marks tokens to which the self-teacher assigns a lower probability
  than the student and green a higher one; the color saturates at
  $|\delta_t| = 2$.}
  \label{fig:guidance_example}
\end{figure}

Figure~\ref{fig:guidance_example} gives an example in which the full
trace provides more informative guidance than the answer. It shows the
token-level guidance under the two forms of context on the same passage
of an unsuccessful rollout of the \texttt{Qwen3-4B} student before training.
In it the student computes the length of an interval incorrectly, writing
$1/5^{m+1}$ where the difference is $4/5^{m+1}$. With the full trace as
context, which contains the correct derivation, the guidance is strongly
negative on the incorrect numerator and on the tokens that carry the
result into the next step. With the ground-truth answer as context, the
guidance on the same tokens is close to zero, since the answer alone does
not indicate which intermediate step is wrong.

\section{Wording of Incorrect and Correct Responses}
\label{app:wording}

The case study in Section~\ref{sec:anatomy} shows a single response. Here we
compare the wording of responses across the three benchmarks, for the
base \texttt{Qwen3-4B} and the student trained with BSD on
\texttt{Qwen3-32B} traces, with four samples per problem from each model.
On 26 problems the base model gives at least one incorrect response and
the BSD student at least one correct response; these comprise 45 incorrect
responses of the base model and 69 correct responses of the BSD student.
For every word $n$-gram with $n \le 3$ that is frequent in these
responses (1{,}508 $n$-grams), we compute its rate per 1{,}000 words in
the two groups for each problem and a paired $t$-statistic over the
problems. The rates of 24 $n$-grams differ with $|t| > 3$, where about
four are expected by chance; Table~\ref{tab:wording} lists examples.
Incorrect responses of the base model contain more expressions of
negation and doubt, and correct responses of the BSD student more
expressions of verification. The correct responses are also shorter,
with 19{,}057 tokens on average against 24{,}279. This comparison changes
the model and the correctness of the response at the same time, so the
differences cannot be attributed to BSD alone. We report them as a
description of the two groups of responses and not as independent
evidence that BSD induces self-correction.

\begin{table}[t]
\begin{center}
\small
\setlength{\tabcolsep}{4.5pt}
\begin{tabular}{lccc}
\toprule
$n$-gram & Base, incorrect & BSD, correct & $t$ \\
\midrule
but this is & 0.84 & 0.38 & $-4.3$ \\
this is not & 0.45 & 0.18 & $-4.2$ \\
not sure & 0.58 & 0.29 & $-3.4$ \\
try to & 1.21 & 0.66 & $-3.4$ \\
approach & 0.80 & 0.58 & $-3.3$ \\
\midrule
which matches & 0.12 & 0.21 & $+4.3$ \\
indeed & 0.12 & 0.27 & $+3.3$ \\
verify & 0.15 & 0.23 & $+3.1$ \\
is indeed & 0.10 & 0.22 & $+3.1$ \\
\bottomrule
\end{tabular}
\end{center}
\caption{Rates per 1{,}000 words of selected $n$-grams in incorrect
responses of the base \texttt{Qwen3-4B} and correct responses of the
student trained with BSD, on the 26 problems that have both. $t$ is the
paired $t$-statistic over the problems.}
\label{tab:wording}
\end{table}

\section{Selected Checkpoints}
\label{app:steps}

Table~\ref{tab:steps} lists the training step of every checkpoint
evaluated in Table~\ref{tab:main}.

\begin{table}[!htb]
\begin{center}
\small
\setlength{\tabcolsep}{3.5pt}
\resizebox{\columnwidth}{!}{%
\begin{tabular}{l ccc ccc ccc}
\toprule
 & \multicolumn{3}{c}{\texttt{Qwen3-32B}} & \multicolumn{3}{c}{\texttt{Nemotron-3-30B-A3B}} & \multicolumn{3}{c}{\texttt{GPT-5.6 Luna}} \\
\cmidrule(lr){2-4} \cmidrule(lr){5-7} \cmidrule(lr){8-10}
Student & SFT & OPD & BSD & SFT & OPD & BSD & SFT & OPD & BSD \\
\midrule
\texttt{Qwen3-4B} & 25 & 75 & 125 & 25 & 125 & 25 & 25 & 75 & 50 \\
\texttt{Qwen3-4B-Instruct} & 25 & 150 & 150 & 50 & 75 & 100 & 25 & 25 & 150 \\
\texttt{Qwen3-1.7B} & 150 & 125 & 125 & 25 & 150 & 50 & 25 & 150 & 25 \\
\texttt{Qwen3-0.6B} & 25 & 50 & 25 & 25 & 150 & 50 & 25 & 100 & 50 \\
\bottomrule
\end{tabular}}
\end{center}
\caption{Training step of the checkpoint selected on the development set
for each run of Table~\ref{tab:main}. Checkpoints are saved every 25
steps. As in Table~\ref{tab:main}, the OPD column is vanilla OPD for
\texttt{Qwen3-32B}, xOPD for \texttt{Nemotron-3-30B-A3B}, and ROPD for \texttt{GPT-5.6 Luna}.}
\label{tab:steps}
\end{table}

\section{Pass@4 Results}
\label{app:matrices}

Table~\ref{tab:pass4} reports Pass@4 for the checkpoints of
Table~\ref{tab:main}.

\begin{table*}[t]
\begin{center}
\small
\setlength{\tabcolsep}{4pt}
\resizebox{\textwidth}{!}{%
\begin{tabular}{l cccc cccc cccc}
\toprule
 & \multicolumn{4}{c}{Shared tokenizer (\texttt{Qwen3-32B})} & \multicolumn{4}{c}{Cross-tokenizer (\texttt{Nemotron-3-30B-A3B})} & \multicolumn{4}{c}{Closed API (\texttt{GPT-5.6 Luna})} \\
\cmidrule(lr){2-5} \cmidrule(lr){6-9} \cmidrule(lr){10-13}
Method & AIME24 & AIME25 & HMMT25 & Avg & AIME24 & AIME25 & HMMT25 & Avg & AIME24 & AIME25 & HMMT25 & Avg \\
\midrule
\multicolumn{13}{l}{\texttt{Qwen3-4B}} \\
Base & 86.7 & 80.0 & 56.7 & 74.4 & 86.7 & 80.0 & 56.7 & 74.4 & 86.7 & 80.0 & 56.7 & 74.4 \\
+ SFT & 80.0 & 76.7 & 56.7 & 71.1 & 83.3 & 80.0 & \textbf{63.3} & \textbf{75.6} & \textbf{83.3} & \textbf{80.0} & 50.0 & 71.1 \\
+ OPD & \textbf{83.3} & \textbf{80.0} & 50.0 & 71.1 & \textbf{86.7} & 80.0 & 60.0 & \textbf{75.6} & \textbf{83.3} & 76.7 & \textbf{56.7} & \textbf{72.2} \\
\rowcolor{bsdrow} + BSD & 80.0 & \textbf{80.0} & \textbf{60.0} & \textbf{73.3} & \textbf{86.7} & \textbf{83.3} & 56.7 & \textbf{75.6} & \textbf{83.3} & 76.7 & 53.3 & 71.1 \\
\midrule
\multicolumn{13}{l}{\texttt{Qwen3-4B-Instruct}} \\
Base & 70.0 & 63.3 & 43.3 & 58.9 & 70.0 & 63.3 & 43.3 & 58.9 & 70.0 & 63.3 & 43.3 & 58.9 \\
+ SFT & 63.3 & \textbf{63.3} & 40.0 & 55.6 & 63.3 & 60.0 & 40.0 & 54.4 & 70.0 & \textbf{73.3} & 43.3 & 62.2 \\
+ OPD & \textbf{70.0} & 60.0 & \textbf{46.7} & \textbf{58.9} & \textbf{66.7} & \textbf{63.3} & 50.0 & 60.0 & \textbf{73.3} & 66.7 & 43.3 & 61.1 \\
\rowcolor{bsdrow} + BSD & \textbf{70.0} & 60.0 & \textbf{46.7} & \textbf{58.9} & \textbf{66.7} & \textbf{63.3} & \textbf{56.7} & \textbf{62.2} & \textbf{73.3} & 66.7 & \textbf{53.3} & \textbf{64.4} \\
\midrule
\multicolumn{13}{l}{\texttt{Qwen3-1.7B}} \\
Base & 73.3 & 46.7 & 33.3 & 51.1 & 73.3 & 46.7 & 33.3 & 51.1 & 73.3 & 46.7 & 33.3 & 51.1 \\
+ SFT & 66.7 & \textbf{53.3} & 36.7 & 52.2 & 73.3 & 50.0 & 40.0 & 54.4 & \textbf{76.7} & \textbf{56.7} & 26.7 & 53.3 \\
+ OPD & \textbf{73.3} & 40.0 & 36.7 & 50.0 & 66.7 & 50.0 & \textbf{46.7} & 54.4 & 66.7 & 53.3 & \textbf{36.7} & 52.2 \\
\rowcolor{bsdrow} + BSD & 70.0 & 50.0 & \textbf{40.0} & \textbf{53.3} & \textbf{76.7} & \textbf{56.7} & 43.3 & \textbf{58.9} & 73.3 & 53.3 & \textbf{36.7} & \textbf{54.4} \\
\midrule
\multicolumn{13}{l}{\texttt{Qwen3-0.6B}} \\
Base & 23.3 & 33.3 & 13.3 & 23.3 & 23.3 & 33.3 & 13.3 & 23.3 & 23.3 & 33.3 & 13.3 & 23.3 \\
+ SFT & 16.7 & 20.0 & 13.3 & 16.7 & 13.3 & 23.3 & \textbf{13.3} & 16.7 & 3.3 & 13.3 & 6.7 & 7.8 \\
+ OPD & 26.7 & \textbf{33.3} & \textbf{23.3} & \textbf{27.8} & \textbf{30.0} & \textbf{30.0} & \textbf{13.3} & \textbf{24.4} & 26.7 & 26.7 & 13.3 & 22.2 \\
\rowcolor{bsdrow} + BSD & \textbf{33.3} & 30.0 & 13.3 & 25.6 & 23.3 & \textbf{30.0} & \textbf{13.3} & 22.2 & \textbf{30.0} & \textbf{33.3} & \textbf{23.3} & \textbf{28.9} \\
\bottomrule
\end{tabular}}
\end{center}
\caption{Pass@4, the share of problems solved by at least one of the four
samples, on each benchmark and averaged over the three (Avg). Checkpoints,
protocol, and layout as in Table~\ref{tab:main}, including the OPD
variant of each teacher; bold marks the best score among the three
methods for each teacher--student pair.}
\label{tab:pass4}
\end{table*}

\begin{figure*}[t]
  \centering
  \small
  \begin{tabular}{@{}c@{\hspace{0.015\textwidth}}c@{\hspace{0.015\textwidth}}c@{}}
  \includegraphics[width=0.32\textwidth]{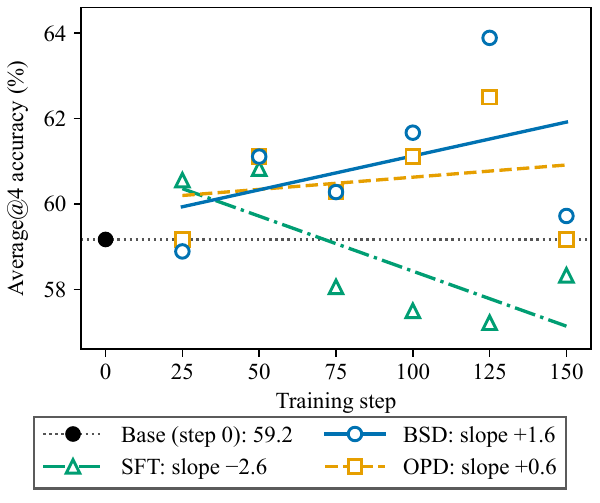} &
  \includegraphics[width=0.32\textwidth]{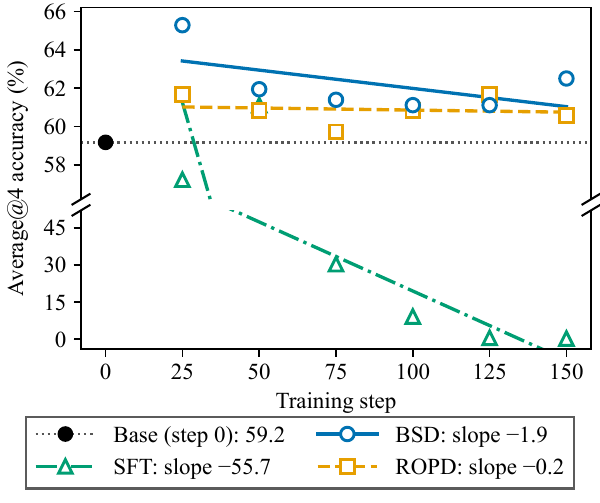} &
  \includegraphics[width=0.32\textwidth]{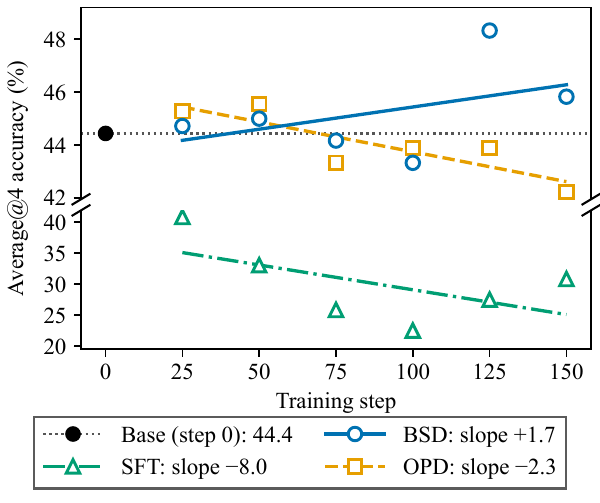} \\
  \shortstack{(a) \texttt{Qwen3-4B} student,\\\texttt{Qwen3-32B} teacher} &
  \shortstack{(b) \texttt{Qwen3-4B} student,\\\texttt{GPT-5.6 Luna} teacher} &
  \shortstack{(c) \texttt{Qwen3-4B-Instruct} student,\\\texttt{Qwen3-32B} teacher} \\
  \end{tabular}
  \caption{Avg@4 across the three benchmarks at every saved checkpoint
  for three further settings. Lines are linear fits over the checkpoints, and the
  legends give their slopes in percentage points per 100 training steps.}
  \label{fig:curves_app}
\end{figure*}

\section{Additional Training Progression Results}
\label{app:dev_curves}

Figure~\ref{fig:curves_app} shows the score trend for three further
settings, in the format of Figure~\ref{fig:progression}d. In each of them the
score of SFT declines over training, while BSD stays stable.

\section{Out-of-Domain Evaluation}
\label{app:ood}

To examine how the distillation methods affect out-of-domain performance, 
we evaluate the \texttt{Qwen3-4B} student on general knowledge and legal questions, 
at every saved checkpoint of BSD, SFT, and the OPD variant of each teacher.
We use two sets of 200 multiple-choice questions: from
MMLU-Pro~\citep{wang2024mmlupro}, 20 questions from each of its ten
categories other than mathematics, physics, chemistry, and engineering,
and from LEXam~\citep{fan2026lexam}, 200 four-choice questions in English
and German, sampled in the proportion of the two languages in the
benchmark. The protocol is that of Table~\ref{tab:eval_config}, and the
base model scores 63.5 on the MMLU-Pro questions and 45.9 on the LEXam
questions.

\begin{figure*}[t]
  \centering
  \includegraphics[width=0.92\textwidth]{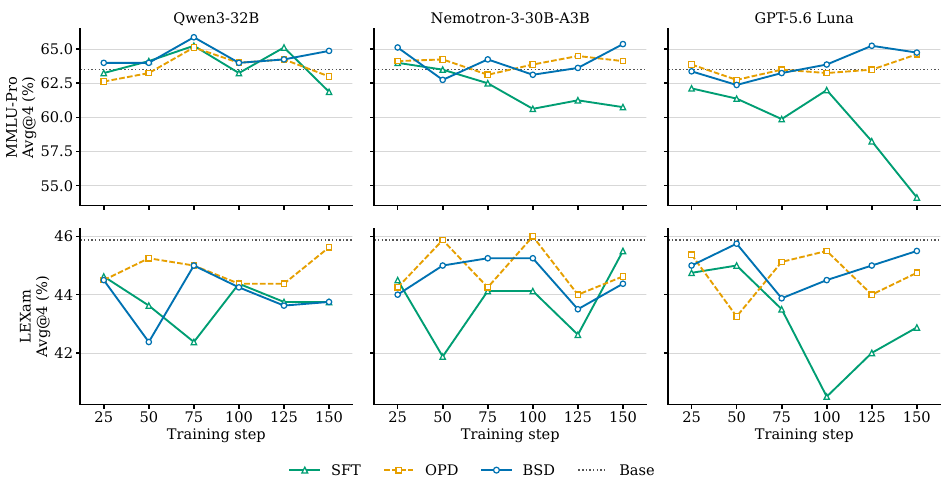}
  \caption{Avg@4 of the \texttt{Qwen3-4B} student on out-of-domain
  questions at every saved checkpoint. Rows are the MMLU-Pro and LEXam
  questions, and columns are the external teachers. OPD is vanilla OPD,
  xOPD, and ROPD for the three teachers, respectively. SFT falls below
  the base model over training with \texttt{Nemotron-3-30B-A3B} and
  \texttt{GPT-5.6 Luna}, whereas BSD and OPD stay at its level.}
  \label{fig:ood}
\end{figure*}

Figure~\ref{fig:ood} shows that SFT is unstable during training,
whereas BSD and the OPD variants are not. On the MMLU-Pro
questions, SFT with \texttt{GPT-5.6 Luna} falls steadily from 62.1 at
step 25 to 54.1 at step 150, 9.4 points below the base model, and SFT with
\texttt{Nemotron-3-30B-A3B} falls from 64.0 to 60.8. On the LEXam
questions, SFT with \texttt{GPT-5.6} falls to 40.5 at step 100, 5.4
points below the base model. Under a bootstrap paired over questions,
these three differences from the base model exclude zero at the 95\%
level, whereas no difference of SFT with \texttt{Qwen3-32B} does. In
contrast, at every checkpoint and for all three teachers, BSD and the
OPD variants are never more than 1.1 points below the base model on the
MMLU-Pro questions and 3.5 points on the LEXam questions. Learning from
the student's own rollouts therefore preserves its abilities outside the
training domain, whereas imitating the teacher's reasoning traces can erase
them.

\begin{figure*}[t]
  \centering
  \includegraphics[width=0.92\textwidth]{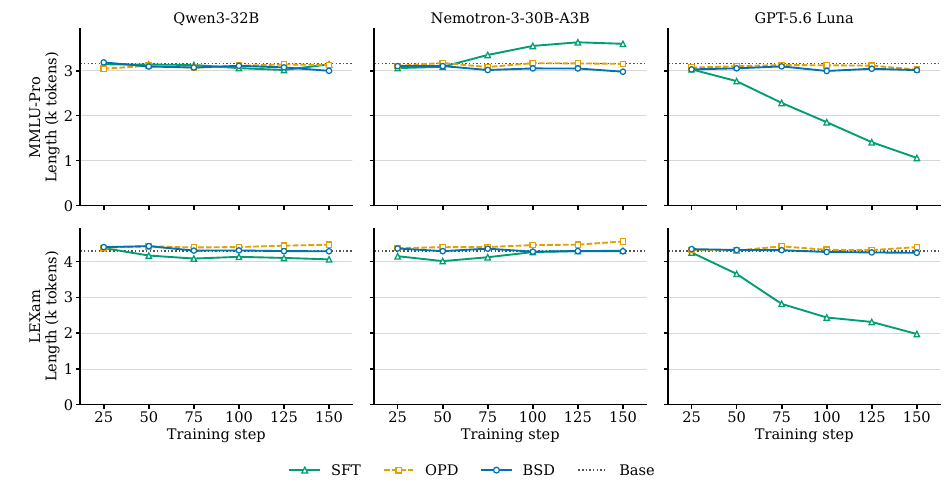}
  \caption{Mean response length of the \texttt{Qwen3-4B} student on the
  out-of-domain questions at every saved checkpoint, in the layout of
  Figure~\ref{fig:ood}. SFT shortens the responses with
  \texttt{GPT-5.6 Luna} and lengthens them on the MMLU-Pro questions with
  \texttt{Nemotron-3-30B-A3B}, whereas BSD and OPD keep the length of the
  base model.}
  \label{fig:ood_length}
\end{figure*}

Figure~\ref{fig:ood_length} shows the mean response length in the same
evaluations. With \texttt{GPT-5.6 Luna}, SFT shortens the responses from
3{,}031 to 1{,}059 tokens on the MMLU-Pro questions and from 4{,}255 to
1{,}975 tokens on the LEXam questions, against 3{,}159 and 4{,}302 tokens
for the base model, and with \texttt{Nemotron-3-30B-A3B} it lengthens
them to 3{,}600 tokens on the MMLU-Pro questions. BSD and the OPD
variants stay within 300 tokens of the base model at every checkpoint.
The loss of out-of-domain accuracy under SFT thus coincides with the
change in response length that it shows on mathematics
(Section~\ref{sec:progression}).

Table~\ref{tab:ood} summarizes each run by the mean and standard
deviation of its score over the six checkpoints. BSD averages 63.8 to
64.5 on the MMLU-Pro questions and 43.9 to 44.9 on the LEXam questions,
with a standard deviation of at most 1.1 points, and the OPD variants
behave similarly. SFT with \texttt{GPT-5.6 Luna} averages 59.6 on the
MMLU-Pro questions with a standard deviation of 3.1 points. The base
model has a single checkpoint, so we report its standard deviation over
the four sampling seeds; on the LEXam questions it is 4.9 points, and
the gap between the base model and the trained students there is
comparable to this sampling noise.

\begin{table}[t]
\begin{center}
\small
\setlength{\tabcolsep}{4pt}
\resizebox{\columnwidth}{!}{%
\begin{tabular}{llcc}
\toprule
Teacher & Method & MMLU-Pro & LEXam \\
\midrule
-- & Base & $63.5 \pm 1.9$ & $45.9 \pm 4.9$ \\
\midrule
\multirow{3}{*}{\texttt{Qwen3-32B}} & SFT & $63.8 \pm 1.3$ & $43.8 \pm 0.8$ \\
 & OPD & $63.7 \pm 0.9$ & $44.9 \pm 0.5$ \\
 & BSD & $64.5 \pm 0.8$ & $43.9 \pm 0.9$ \\
\midrule
\multirow{3}{*}{\texttt{Nemotron-3-30B-A3B}} & SFT & $62.1 \pm 1.5$ & $43.8 \pm 1.3$ \\
 & xOPD & $64.0 \pm 0.5$ & $44.8 \pm 0.9$ \\
 & BSD & $64.0 \pm 1.1$ & $44.6 \pm 0.7$ \\
\midrule
\multirow{3}{*}{\texttt{GPT-5.6 Luna}} & SFT & $59.6 \pm 3.1$ & $43.1 \pm 1.7$ \\
 & ROPD & $63.6 \pm 0.6$ & $44.7 \pm 0.9$ \\
 & BSD & $63.8 \pm 1.1$ & $44.9 \pm 0.7$ \\
\bottomrule
\end{tabular}}
\end{center}
\caption{Mean and standard deviation of Avg@4 over the six saved
checkpoints of each run of the \texttt{Qwen3-4B} student on the
out-of-domain questions of Figure~\ref{fig:ood}. For the base
model, the standard deviation is over its four sampling seeds.}
\label{tab:ood}
\end{table}

\end{document}